\PassOptionsToPackage{table}{xcolor}
\documentclass{article} 
\usepackage{iclr2027_conference,times}

\usepackage{amsmath,amsfonts,bm}

\def\eqref#1{equation~\ref{#1}}

\def\1{\bm{1}}

\DeclareMathAlphabet{\mathsfit}{\encodingdefault}{\sfdefault}{m}{sl}
\SetMathAlphabet{\mathsfit}{bold}{\encodingdefault}{\sfdefault}{bx}{n}

\usepackage{xcolor}

\iclrfinalcopy   
\usepackage{amsthm}
\theoremstyle{plain}
\newtheorem{theorem}{Theorem}

\newtheorem{corollary}[theorem]{Corollary}
\newtheorem{proposition}[theorem]{Proposition}
\theoremstyle{definition}
\newtheorem{assumption}{Assumption}
\theoremstyle{remark}
\newtheorem{remark}{Remark}
\usepackage{hyperref}
\usepackage{url}
\usepackage{booktabs}
\usepackage{enumitem}
\usepackage{pifont}
\usepackage{booktabs}
\usepackage{graphicx}
\usepackage{multirow}
\usepackage{multicol}
\usepackage[table]{xcolor}
\usepackage{enumitem}
\usepackage{adjustbox}
\usepackage{capt-of} 
\usepackage{subcaption} 

\usepackage{pgfplots}
\usepackage{pgfplotstable}

\usepgfplotslibrary{groupplots} \usetikzlibrary{calc}

\pgfplotsset{compat=1.18}
\definecolor{darkblue}{RGB}{0, 0, 139}
\definecolor{darkred}{RGB}{139, 0, 0}

\newcommand{\dr}[1]{\textcolor{darkred}{#1}}
\newcommand{\db}[1]{\textcolor{darkblue}{#1}}

\newcommand{\leanfix}[1]{{#1}}

\definecolor{revred}{RGB}{215,0,0}
\definecolor{revblue}{RGB}{0,0,225}
\definecolor{revbright}{RGB}{0,150,255}
\definecolor{revpurple}{RGB}{140,0,200}

\newcommand{\gram}[1]{{#1}}
\newcommand{\filled}[1]{{#1}}
\newcommand{\aifix}[1]{{#1}}

\newcommand{\ours}{\textsc{WebMix}}
\newcommand{\tl}{\textsc{T2L}}

\newcommand{\best}[1]{\cellcolor{teal!10}\textbf{#1}}

\title{Don't Retrain, Remix: Adaptation of Web Agents using Mixture Hypernetworks}

\author{Md Farhan Ishmam, Minh Pham-Dinh, Kenneth Marino\\
 Kahlert School of Computing, University of Utah\\
\texttt{\{farhan.ishmam,m.pham-dinh,kenneth.marino\}@utah.edu} \\
}

\begin{document}

\maketitle
\lhead{Preprint} 

\vspace{-24pt} 
\begin{center}
\url{https://webmix-hyper.github.io/}
\end{center}

\begin{abstract}
The web is not static or uniform. Websites that serve the same purpose can look and navigate very differently. Even within a single site, successive redesigns can change how familiar
\gram{features are presented and accessed.} Humans adapt to this by recognizing familiar interaction patterns and reusing what they already know. We give web agents a similar ability with \ours{}, a mixture hypernetwork that
\gram{adapts them to unseen interfaces by composing previously trained low-rank adapters.} \ours{} begins with a bank of adapters trained on source interfaces. At deployment, a scout agent explores the target interface and summarizes its observations into a compact manual. A lightweight router uses this manual to predict mixing coefficients over the adapter bank, which combine the existing adapters into a new one for inference. Adapting to a new interface takes only a scouting run and a forward pass, without gradient updates on the target.
\gram{With Qwen and Llama models, we evaluate \textit{domain transfer} on TimeWarp and WebArena-Lite-v2, and \textit{version transfer} on TimeWarp.}
In all three settings, \gram{\ours{} matches} or outperforms in-context learning and LoRA baselines, often matching the average success of specialists trained directly on those interfaces.
\aifix{On TimeWarp domain transfer, it gains about 8 percentage points on average over the best baseline with both models.}
These gains do not come at the cost of seen interfaces, where \ours{} keeps its lead{, and extend to version transfer}.
\aifix{Adapting to a new interface can therefore rely on exploration and composition instead of retraining.}

\end{abstract}

\begin{figure}[h]
    \centering
    \includegraphics[width=\linewidth]{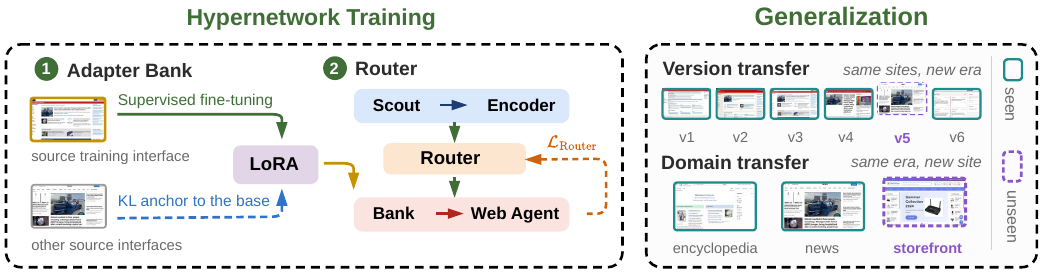}
    \caption{\filled{\textbf{\ours{} at a glance.} We train a bank of interface-specific LoRA adapters, each KL-anchored to the base model on the other source interfaces. At deployment, a scout explores the unseen interface and writes a short manual. A router reads the encoded manual and mixes the bank into one adapter for the web agent. We test transfer to unseen domains and unseen versions.}}
    \label{fig:teaser}
\end{figure}

\section{Introduction}
\label{sec:intro}
Humans compose and transfer skills to unfamiliar interfaces with little effort. A person who has booked flights on American Airlines' website can book the same flight through their credit card's travel portal, following the same search, filter, and checkout flow in a layout they have never seen. The same holds across time: when a site is redesigned, its regular users are slowed down for a few visits, not retaught. A fine-tuned web agent can do neither.
\gram{Mind2Web~\citep{deng2023mind2web} reports a steady drop in performance from its cross-task split to its cross-website and cross-domain splits. TimeWarp~\citep{ishmam2026timewarp}, which renders the same sites in the styles of six eras of web design, shows that an agent trained on one era degrades on the next.}
Frontier models reduce this brittleness through scale~\gram{\citep{Zheng2024GPT4VisionIA, He2024WebVoyagerBA, anthropic_cua, openai_cua, browser_use2024}}, but their closed weights and serving cost raise privacy and deployment concerns~\gram{\citep{Qi2024WebRLTL, Zhang2025PrivWebUA}}.

Small open-weight models that run on commodity hardware are the practical alternative~\citep{grattafiori2024llama3herdmodels, qwen3.5}, and this brittleness hits them hardest. They become competent web agents only after fine-tuning on demonstrations, typically through a low-rank adapter (LoRA)~\citep{hu2022lora} on a frozen base, and every new site or redesign calls for a new corpus and a new adapter. Their performance is therefore bounded by the demonstrations they are trained on, and web demonstrations are expensive to collect: each one requires a human operating a live site, and the site may be redesigned before the corpus is large enough.
\aifix{We therefore ask whether an agent can get more out of the demonstrations it already has.}
Facebook has altered its interface repeatedly since 2004, but it has always had a search bar. Amazon and Wikipedia serve different purposes entirely, yet browsing a
\gram{catalog} of products and browsing an \gram{encyclopedia} of articles are the same operation: filtering, paging, and selecting from a list. If an adapter trained on one interface has already learned these operations, that knowledge ought to be transferable to the next.

\aifix{Recent work adapts small web agents with less supervision. It reduces} target-interface demonstrations to one or two examples~\citep{verma-etal-2025-adaptagent}, \gram{synthesizes} trajectories through autonomous exploration of the target~\citep{Murty2024NNetNavUL}, \gram{transfers} reusable action routines across websites~\citep{Wang2024AgentWM}, \gram{or learns} lightweight alignment vectors from deployment-time interaction~\citep{Chen2025GroundedTA}. All of these still treat each new interface as a separate adaptation problem.
\aifix{They reuse knowledge about websites, but none of them learns how to adapt, so the next interface starts from scratch again. A hypernetwork can learn such an adaptation rule, because it generates an adapter from a description of the target~\citep{ha2017hypernetworks, charakorn2025t2l}. For web interfaces, however, generating adapter weights directly is hard.}
A family of interfaces provides only a handful of conditionings from which to regress millions of weights, and LoRA factors are not identifiable, so adapters that behave alike can lie far apart in weight space~\citep{charakorn2025t2l}.

We introduce \ours{}, a mixture hypernetwork~\gram{\citep{qu2023hmoehypernetworkbasedmixtureexperts}} that learns to adapt web agents by composing LoRA adapters trained on familiar interfaces. A scout briefly explores a new interface and summarizes its observations in a compact manual. Conditioned on this description, a small router predicts one mixing coefficient per adapter and combines their weight updates into a single adapter for the frozen base policy.
\gram{We train the router on source demonstrations through the composed policy and keep the adapter bank fixed. Adaptation then requires only scouting and one forward pass, with no further gradient updates.}

On TimeWarp~\gram{\citep{ishmam2026timewarp}}, \ours{} achieves 56.1\% success on the unseen News domain with Qwen-3.5 4B, exceeding transferred LoRA by 10.6 percentage points. Across held-out domains, average gains reach 8.3 points for Qwen and 7.5 points for Llama-3.1 8B, nearly doubling Llama's success and matching the average performance of target-trained specialists for both models. Improvements extend to held-out TimeWarp versions and WebArena-Lite-v2 domains, where Qwen gains 3.2 and 4.6 points, respectively. The benefits also extend to seen interfaces: with target domains included in training, Qwen achieves 42.4\% overall success on WebArena-Lite-v2, outperforming LoRA trained on all domains by 4.3 points.

\aifix{Our contributions are the following:}
\begin{itemize}[leftmargin=20pt]
    \item[\ding{69}] We introduce \ours{}, a mixture hypernetwork that learns to compose existing LoRA adapters from a scouted interface description. The learned adaptation rule transfers to new websites and redesigns without parameter optimization on the target interface \filled{(\S\ref{sec:methodology})}.

    \item[\ding{69}] We evaluate domain transfer on TimeWarp and WebArena-Lite-v2 and version transfer on TimeWarp with two \gram{open-weight models (\S\ref{sec:experiments})}. {On held-out TimeWarp domains, \ours{} matches target-trained specialists on average for both models, reaches 56.1\% success on unseen News with Qwen (+10.6 percentage points over transferred LoRA), and nearly doubles Llama's average domain-transfer success.}

    \item[\ding{69}] We decompose held-out imitation risk into bank approximation and routing terms, clarifying the limits of adapter composition \filled{(\S~\ref{sec:theory})}. Ablations compare composition with direct adapter generation and examine how scouting and adapter regularization contribute to transfer \filled{(\S\ref{subsec:ResultsAndAnalysis})}.
\end{itemize}

\section{Related Work}
\label{sec:related}
\paragraph{Web Agents and Interface Generalization.}
\gram{Web-agent benchmarks test generalization to unseen interfaces along different axes of variation.} Along the cross-site axis, Mind2Web evaluates agents on static demonstrations split by task, website, and domain \citep{deng2023mind2web}, while WebArena and WorkArena provide self-hosted, fully functional websites \citep{zhou2024webarena,drouin2024workarena}, and WebVoyager and Online-Mind2Web move evaluation to live websites \citep{He2024WebVoyagerBA,Xue2025AnIO}. Along the temporal axis, TimeWarp evaluates agents on successive eras of web design \citep{ishmam2026timewarp}. Complementary diagnostic suites isolate individual factors by perturbing the layout, interaction semantics, visual attributes, or network events of otherwise fixed tasks \citep{bai2026stresswebdiagnosticbenchmarkweb, Kara2025WAREXWA}.
\aifix{On all of these benchmarks, agents lose accuracy on interfaces they were not trained on.}
{Mind2Web reports substantially lower performance for fine-tuned agents on unseen websites and domains than on its cross-task split \citep{deng2023mind2web}, and TimeWarp finds that text-based agents become brittle under interface changes after fine-tuning on a single version \citep{ishmam2026timewarp}.}
A practical web agent should therefore adapt at test time to visually and functionally unfamiliar interfaces without relying on extensive demonstrations.
\gram{We evaluate \ours{} along both axes. TimeWarp provides both domain shifts and version shifts, and WebArena-Lite-v2 \citep{liu2024visualagentbench, liu2026scalecua} provides domain shifts.}

\paragraph{Adapting Web Agents to Unseen Interfaces.}
An agent facing an unfamiliar interface must learn something about it before it can act well. Prior work differs in where that knowledge comes from and where it is kept. The most direct source is a human: AdaptAgent adapts multimodal agents to
new websites from up to two human demonstrations, using in-context demonstrations for proprietary models and meta-learned adaptation for open-weight models \citep{verma-etal-2025-adaptagent}. Demonstrations, however, are costly to collect, so a second line of work lets the agent gather its own experience through guided or unguided exploration \citep{Murty2024NNetNavUL,Su2025LearnbyinteractAD,Pahuja2025ExplorerSE,Sun2024OSGenesisAG}. Beyond demonstrations and raw experience, an agent can also carry forward explicit knowledge. Non-parametric methods keep the weights fixed and store what the agent learns outside them: as reusable workflows \gram{\citep{Wang2024AgentWM}} or episodic memories \gram{\citep{sarch2025vlmagentsgeneratememories}}, as an instruction manual written through interaction \citep{chen2024automanualconstructinginstructionmanuals}, as skills distilled from practice \citep{Zheng2025SkillWeaverWA,wang2026webxskillskilllearningautonomous}, or as summaries of environment dynamics learned at deployment time \citep{Chen2025GroundedTA}.
Like AutoManual \gram{\citep{chen2024automanualconstructinginstructionmanuals}}, \ours{} constructs an explicit description of the environment from interaction. \gram{AutoManual uses this description to condition planning, whereas we use it to predict a composition of existing adapters.} \aifix{More generally, prior methods treat each new interface as a separate adaptation problem. \ours{} instead learns an interface-conditioned rule for composing adapters on the source interfaces and applies it unchanged to the target interface.} Target-side exploration supplies the conditioning information, without target demonstrations or test-time gradient updates.

\paragraph{Adapter Composition and Generation.}
Low-rank adaptation \citep{hu2022lora} \gram{lets many adapters share} one frozen base model. \aifix{A growing body of work reuses such collections instead of training a new model for each task \citep{Yadav2024WhatMF}.} The simplest form of reuse combines adapters in weight space. Merging methods add or average updates arithmetically \citep{ilharco2023taskarithmetic,yadav2023ties,wortsman2022soups}. Other approaches optimize composition coefficients on target examples, as in LoraHub \citep{huang2024lorahub}, or compose adapters dynamically
during inference through layer-wise gating in MoLE \citep{wu2024mole} and token-level routing in Arrow and PHATGOOSE \citep{ostapenko2024arrow,muqeeth2024phatgoose}. A more ambitious alternative skips the bank altogether and generates the adapter with a hypernetwork \citep{ha2017hypernetworks}, conditioned on contexts and documents \citep{charakorn2026doctoloralearninginstantlyinternalize}, task descriptions and prompts \citep{charakorn2025t2l,liang2025dnd}, whole code repositories \citep{hotsko2026code2lorahypernetworkgeneratedadapterscode}, or per-sample memories in continual learning \citep{wang2026hylovqadynamichypernetworkgeneratedlowrank}. Generation is flexible, but it is hard to learn. The generator must {emit} every entry of every LoRA factor, and because LoRA factors are not identifiable, many different factors encode the same update. This is manageable with abundant conditionings but not with few: Drag-and-Drop LLMs, for example, improve little over their training adapters when trained on only a few datasets \citep{liang2025dnd}, and a family of web interfaces provides only a handful.
\gram{\ours{} sits between these two lines. Like a hypernetwork, it maps a description of the target to an adapter in one forward pass. Like merging, it builds that adapter from a frozen bank. It predicts only one coefficient per bank adapter and emits a single standard LoRA adapter that an unmodified inference stack can serve.}
We compare against Text-to-LoRA conditioned on the same interface representation, following \citet{hotsko2026code2lorahypernetworkgeneratedadapterscode}{, and analyze {an idealization of it} in the {Appendix}~\S\ref{sec:theory:twin}}.

%
\begin{figure}
    \centering
    \includegraphics[width=\linewidth]{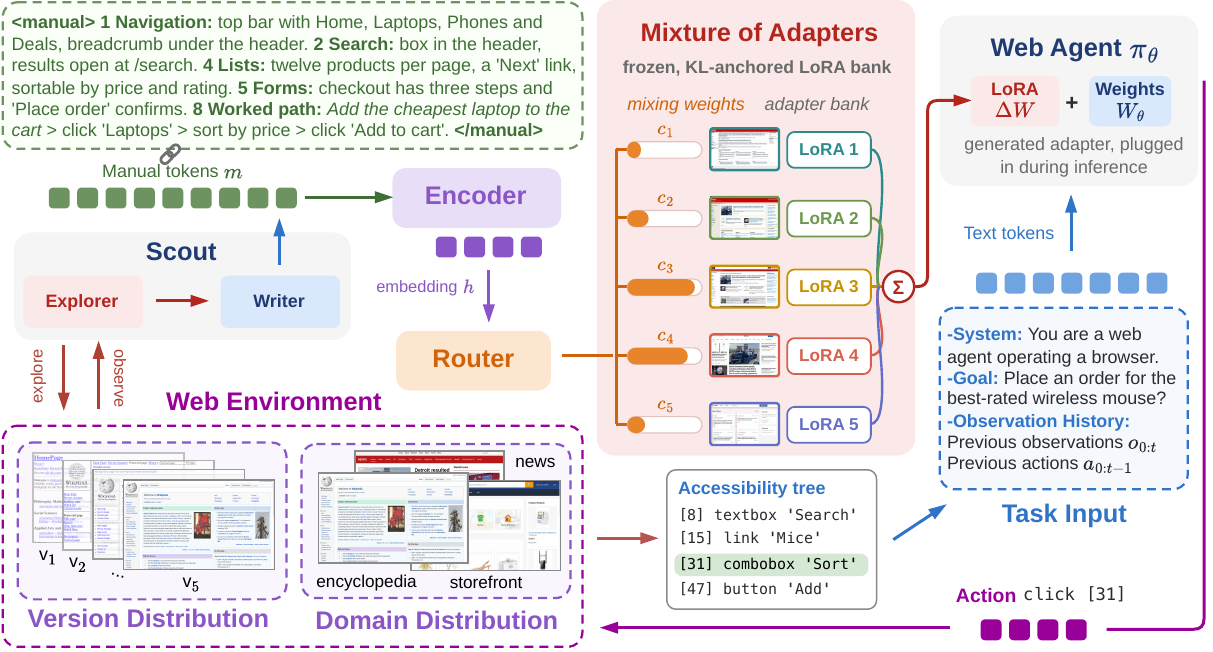}
    \caption{\textbf{\gram{Overview of \ours{}.}}
A scout explores the target website. Its Explorer interacts with the environment, and its Writer turns the observations into a compact manual $m$.
An encoder maps $m$ to an embedding $h$, from which a router predicts mixing weights $c_1, \dots, c_5$ over a frozen bank of KL-anchored LoRA adapters.
Their weighted sum is a site-specific adapter $\Delta W$, added to the base weights $W_\theta$ of the agent $\pi_\theta$, which reads the accessibility tree and emits grounded actions (e.g.,~\texttt{click [31]}).
Websites vary across versions ($v_1$--$v_5$) and domains.}
\label{fig:overview}
\end{figure}
\section{\ours{}: Adaptation through Composition}
\label{sec:methodology}

\gram{Figure~\ref{fig:overview} gives an overview. \S\ref{sec:method:setting} states the problem, \S\ref{sec:mixtureHypernetwork} describes the mixture hypernetwork, and \S\ref{sec:method:training} describes its training and its adaptation to an unseen interface.}

\subsection{\gram{Preliminaries}}
\label{sec:method:setting}
A web agent is a language model that completes a task on a website through a browser. It reads a natural-language goal and the current page, and it emits one browser action at a time, such as a click, a typed string, or a scroll, until the task ends. A task succeeds if a verifier accepts the final state of the website. A web environment consists of several \emph{interfaces}. Each interface is either a website domain, such as a wiki or an online shop, or a version, \textit{i.e.}, a design of the website. Interfaces differ in presentation, navigation structure, and content.
{Web agents are often trained using parameter-efficient methods, such as Low Rank Adaptation (LoRA) \citep{hu2022lora}. We now formalize the web environment, agents, LoRA, and our problem setting in the following paragraphs.}

\paragraph{Web environment and agents.} We model the agent's interaction with a web interface as a partially observable Markov decision process (POMDP) $\langle \mathcal{S}, \mathcal{A}, \mathcal{O}, \mathbb{T}, \mathbb{T}_0, \mathbb{O}, R_g\rangle$ \citep{kaelbling1998pomdp}. A state $s\in\mathcal{S}$ comprises the website's front-end (external) and back-end (internal) states. An episode starts from $s_0\sim\mathbb{T}_0$, and browser actions $a\in\mathcal{A}$ change the state according to $s_{t+1}\sim\mathbb{T}(\cdot\mid s_t,a_t)$. The agent receives only an observation $o_t\sim\mathbb{O}(\cdot\mid s_t)\in\mathcal{O}$, in the form of an accessibility tree \citep{zhou2024webarena}. A task has a natural-language goal $g$ and a verifier $V_g:\mathcal{S}\rightarrow\{0,1\}$, which defines the terminal reward $R_g(s_T)=V_g(s_T)$ at the final step $T$. The agent is a policy $\pi_\theta$ with parameters $\theta$ that selects each action from the interaction history, $a_t\sim\pi_\theta(\cdot\mid H_t)$, where $H_t=(g,o_{0:t}, a_{0:t-1})$.
Let $\mathcal{E}$ be the set of interfaces of the environment. We index the source interfaces as $\mathcal{E}_{\mathrm{src}}=\{v_1,\dots,v_K\}\subset\mathcal{E}$, so $K=|\mathcal{E}_{\mathrm{src}}|$, and write an unseen interface as $u\in\mathcal{E}\setminus\mathcal{E}_{\mathrm{src}}$.
\filled{Each source interface $v$ has a demonstration corpus $\mathcal{D}_v=\{(x,y)\}$, and we write $\mathcal{D}_k=\mathcal{D}_{v_k}$. Each input $x$ is a history $H_t$, rendered as a prompt that contains the goal, the accessibility tree of the current page, and a record of the earlier steps. Each target $y=(y_1,\dots,y_{|y|})$ is the expert's response, a token sequence that contains its reasoning and one browser action.}

\paragraph{\gram{Low-rank} adaptation (LoRA).} Fine-tuning all of $\theta$ for each interface would store one copy of the model per interface. Low-rank adaptation (LoRA) \citep{hu2022lora} keeps $\theta$ frozen and learns a low-rank update to selected weight matrices. Let $\Omega$ index the adapted weight matrices layer-wise, and let $W_{\theta,\ell}\in\mathbb{R}^{d_\ell^{\mathrm{out}}\times d_\ell^{\mathrm{in}}}$ be the frozen matrix at $\ell\in\Omega$. LoRA replaces it with
\begin{equation}
    W_\ell = W_{\theta,\ell} + \Delta W_\ell,
    \qquad
    \Delta W_\ell = \frac{\alpha}{r}\,B_\ell A_\ell,
    \label{eq:lora}
\end{equation}
where $A_\ell\in\mathbb{R}^{r\times d_\ell^{\mathrm{in}}}$ and $B_\ell\in\mathbb{R}^{d_\ell^{\mathrm{out}}\times r}$ are trainable factors, $r$ is the adaptation rank, and the scale $\alpha$ sets the size of the update independently of $r$. $A_\ell$ is initialized randomly and $B_\ell$ is set to zero, so training starts from the base policy. We call the collection of updates $\Delta\theta=(\Delta W_\ell)_{\ell\in\Omega}$ an \emph{adapter}, and write $\pi_{\theta+\Delta\theta}$ for the policy with the adapter installed. Adapters can be held separate from the base weights and applied at serving time, so many adapters can share one frozen base model.
\filled{We adapt the attention projections (query, key, value, output) and the MLP projections (gate, up, down), with $r=16$ and $\alpha=32$.}

\paragraph{Problem setting.} We start from a frozen base model and a set of \emph{source} interfaces, each with a corpus of expert demonstrations. Our goal is for the agent to succeed on an \emph{unseen} interface. In \emph{domain transfer}, it is an unseen domain, and in \emph{version transfer}, it is an unseen version. The unseen interface has no demonstrations, and no parameter is optimized on it. Before the agent starts its tasks, it may only make a brief, bounded visit to the unseen interface, which we call \emph{scouting}. We ask whether the parameters the agent needs on the unseen interface can be assembled from what it learned on the source interfaces, using this visit alone.

\subsection{The Mixture Hypernetwork}
\label{sec:mixtureHypernetwork}

A hypernetwork \citep{ha2017hypernetworks} is a network that outputs the parameters of another network. \ours{} uses a hypernetwork $F_\phi$ with trainable parameters $\phi$ to produce the adapter for an interface from a bounded exploration record of it. For an unseen interface $u$ with record $E_u$, $F_\phi$ outputs one update per adapted matrix, which is added to the frozen base weights,
\begin{equation}
    \{\Delta W_{u,\ell}\}_{\ell\in\Omega} = F_\phi(E_u),
    \qquad
    W_{u,\ell} = W_{\theta,\ell} + \Delta W_{u,\ell}.
    \label{eq:setting}
\end{equation}
The updates $\Delta\theta_u=(\Delta W_{u,\ell})_{\ell\in\Omega}$ produce the adapted policy $\pi_{\theta+\Delta\theta_u}$, with all other parameters unchanged.
{The objective is to maximize task success on $u$ without target demonstrations or parameter optimization on the target interface.}
\gram{$F_\phi$ has three parts, described below: a frozen bank of adapters, a small router with the trainable parameters $\phi$, and a frozen scout that maps the record $E_v$ of an interface $v$ to a representation $h_v\in\mathbb{R}^{d_c}$.}

\paragraph{The adapter bank.} The bank holds one LoRA adapter per source interface. For each $v_k$, the bank contains factors $\{(A^{(k)}_\ell,B^{(k)}_\ell)\}_{\ell\in\Omega}$ of rank $r$ and scale $\alpha$, with updates $\Delta W^{(k)}_\ell=(\alpha/r)\,B^{(k)}_\ell A^{(k)}_\ell$. We write $\Delta\theta^{(k)}=(\Delta W^{(k)}_\ell)_{\ell\in\Omega}$ for adapter $k$, and $\pi^{(k)}=\pi_{\theta+\Delta\theta^{(k)}}$ for the policy using it.
{The bank $\{\Delta\theta^{(1)},\dots,\Delta\theta^{(K)}\}$ is the source knowledge that \ours{} transfers to an unseen interface.}
All bank adapters share the base model, the index set $\Omega$, the rank $r$, and the scale $\alpha$, so they can be combined matrix-wise.
{The bank is trained once (Stage~1, \S\ref{subsection:AdapterBank}) and stays frozen afterwards.}

\paragraph{What the hypernetwork predicts.}
A direct hypernetwork, such as Text-to-LoRA \citep{charakorn2025t2l}, predicts every entry of every LoRA factor. The router of \ours{} predicts only $K$ mixing coefficients, one per bank adapter. It is a two-layer perceptron with parameters $\phi=(U_1,b_1,U_2,b_2)$ that maps $h_v$ to a vector of mixture weights over the $K$ bank adapters,
\begin{equation}
    c_v = \operatorname{softmax}\!\left(U_2\,\operatorname{SiLU}(U_1h_v+b_1)+b_2\right)\in\mathbb{R}^K,
    \label{eq:coeff}
\end{equation}
where $U_1\in\mathbb{R}^{d_h\times d_c}$, $b_1\in\mathbb{R}^{d_h}$, $U_2\in\mathbb{R}^{K\times d_h}$, $b_2\in\mathbb{R}^{K}$, and $d_h$ is the hidden width\filled{ ($d_h=256$, Appendix~\ref{subsec:routerDetails})}. The coefficient $c_{v,k}$ is the weight of bank adapter $k$ in the adapter for $v$. A single coefficient vector is shared across all adapted matrices. We initialize $U_2$ and $b_2$ to zero, so the untrained router weights all bank adapters equally.

\paragraph{How the bank is combined.}
The adapter for interface $v$ is the weighted sum of the bank's updates,
\begin{equation}
    \Delta W_{v,\ell} = \sum_{k=1}^{K}c_{v,k}\,\Delta W^{(k)}_\ell.
    \label{eq:exact}
\end{equation}
We compute this sum by concatenating the bank factors along their rank dimension,
\begin{equation}
    A_{v,\ell} =
    \begin{bmatrix}
        c_{v,1}A^{(1)}_\ell \\
        \vdots \\
        c_{v,K}A^{(K)}_\ell
    \end{bmatrix},
    \qquad
    B_{v,\ell} =
    \begin{bmatrix}
        B^{(1)}_\ell & \cdots & B^{(K)}_\ell
    \end{bmatrix}.
    \label{eq:delta}
\end{equation}
The product satisfies $B_{v,\ell}A_{v,\ell}=\sum_k c_{v,k}B^{(k)}_\ell A^{(k)}_\ell$. Averaging the two factors separately would instead weight each adapter's own term by $c_{v,k}^2$ and add cross terms $B^{(j)}_\ell A^{(k)}_\ell$ with $j\neq k$, which by Eq.~\ref{eq:bg:gauge} \gram{(Appendix~\ref{sec:bg:lora})} depends on how each adapter happens to be factored. The composed factors have inner dimension $r_{\mathrm{out}}=Kr$, and we set $\alpha_{\mathrm{out}}=K\alpha$, so that $\alpha_{\mathrm{out}}/r_{\mathrm{out}}=\alpha/r$. The result is a standard LoRA adapter of rank at most $Kr$ and scale $\alpha_{\mathrm{out}}$, and installing it through Eq.~\ref{eq:setting} adds the same update as Eq.~\ref{eq:exact} to the base weights. It is served by an unmodified LoRA inference stack.

\paragraph{The scout.} The scout turns a bounded visit to an interface $v$ into the representation $h_v$ that the router reads. It has three parts: an explorer collects an exploration record $E_v$, a writer summarizes $E_v$ into a text \emph{interface manual} $m_v$, and a \gram{frozen} sentence encoder \gram{$\psi$} embeds $m_v$. None of the three is trained at any stage.
\gram{Appendix~\ref{subsec:scoutDetails} describes each part.} The scout returns two objects for an interface $v$. The first is the manual $m_v$, a readable text summary of the interface. The second is the conditioning vector, the normalized encoding of the manual,
\begin{equation}
    h_v = \frac{\psi(m_v)}{\|\psi(m_v)\|_2}\in\mathbb{R}^{d_c}.
    \label{eq:scout}
\end{equation}
Only $h_v$ enters the hypernetwork, where the router maps it to the mixing coefficients $c_v$ of Eq.~\ref{eq:coeff}.

\subsection{Training and Adaptation}
\label{sec:method:training}

{\ours{} assembles the parameters from low-rank adapters. It trains one adapter per source interface, keeps the adapters in a frozen bank, and learns a small network, the router, that decides how much of each adapter to use on a given interface.} Learning and adaptation proceed in three stages:
\begin{enumerate}[leftmargin=*,itemsep=0pt,topsep=2pt]
    \item \textbf{Stage 1: Adapter Bank Training (\S\ref{subsection:AdapterBank}).} \gram{For each source interface, we fine-tune one adapter on its demonstrations and freeze it in the bank.}
    \item \textbf{Stage 2: Router Training (\S\ref{subsec:routerTraining}).} \gram{The router learns to map the scout's encoded manual of each source interface to mixing weights over the bank, by imitating its demonstrations through the frozen base model and bank.}
    \item \textbf{Stage 3: Adaptation (\S\ref{subsec:adaptation}).} \gram{The scout writes a manual of the unseen interface, and the router mixes the bank into the agent's adapter for it in one forward pass.}
\end{enumerate}
\gram{Both training stages use only source-interface data, so the scouting visit is the only information from the unseen interface that \ours{} uses. Adding a source interface later adds one adapter to the bank and retrains the router, leaving the existing adapters unchanged.}

\subsubsection{Stage 1: Adapter Bank Training}
\label{subsection:AdapterBank}
Each adapter $k$ is trained \gram{with supervised fine-tuning (SFT)} to reproduce the expert on its own interface. An anchoring term also limits how far the adapter moves the base policy on the other source interfaces. We call $\mathcal{D}_k$ the adapter's \emph{own corpus}, and $\mathcal{D}_{\setminus k}=\bigcup_{j\neq k}\mathcal{D}_j$ its \emph{anchor corpus}, which pools the demonstrations of the other $K-1$ source interfaces. Neither corpus contains data from an unseen interface. We minimize the objective
\begin{equation}
    \mathcal{L}_{\mathrm{bank}}^{(k)}
    = \underbrace{\mathbb{E}_{(x,y)\sim\mathcal{D}_k}
    \left[-\frac{1}{|y|}\sum_{i=1}^{|y|}
    \log \pi^{(k)}(y_i\mid x,y_{<i})\right]}_{\text{SFT on training interface}}
    + \beta\,\underbrace{\mathbb{E}_{(x,y)\sim\mathcal{D}_{\setminus k}}
    \left[\frac{1}{|y|}\sum_{i=1}^{|y|}\kappa^{(k)}_{i}\right]}_{\text{anchor on other source interfaces}},
    \label{eq:anchor}
\end{equation}
where $\beta\geq 0$ controls the anchoring strength and $\beta=0$ recovers plain SFT.
\gram{The first term is the SFT loss on the adapter's own interface. The second term penalizes changes to the base policy's likelihood of the demonstrated tokens of the other source interfaces, where $\kappa^{(k)}_i$ is a clipped estimate of the KL divergence (Appendix~\ref{subsec:klAnchor}).}

\subsubsection{Stage 2: Training the Router}
\label{subsec:routerTraining}
Stage 2 trains the router with the bank frozen. Before training, the scout visits each source interface $v_k$ once and returns its representation $h_{v_k}$. From $h_{v_k}$, Eqs.~\ref{eq:coeff} and~\ref{eq:exact} construct the composed update $\Delta\theta_\phi(h_{v_k})=(\Delta W_{v_k,\ell})_{\ell\in\Omega}$. Let $\pi_{\phi,k}=\pi_{\theta+\Delta\theta_\phi(h_{v_k})}$ denote the resulting policy.
We train the router with the \gram{behavior-cloning (BC)} loss averaged over all response tokens of the pooled source corpus,
\begin{equation}
    \mathcal{L}_{\mathrm{router}}(\phi)
    =
    -\frac{1}{N}\sum_{k=1}^{K}\sum_{(x,y)\in\mathcal{D}_k}\sum_{i=1}^{|y|}
    \log \pi_{\phi,k}(y_i\mid x,y_{<i}),
    \qquad
    N=\sum_{k=1}^{K}\sum_{(x,y)\in\mathcal{D}_k}|y|.
\label{eq:obj}
\end{equation}
Prompt tokens are masked, so only the expert's response tokens contribute. Gradients reach the router through the composed updates and the frozen base model, so the router learns which mixtures reproduce the expert on each source interface.
\aifix{Only $\phi$ is trained.} The base model, the bank, and the scout stay frozen, and the representations $h_{v_k}$ are fixed inputs.

\subsubsection{Stage 3: Adaptation}
\label{subsec:adaptation}
In Stage 3, the scout visits the unseen interface $u$ once, with the same procedure and budget as for the source interfaces, and produces $m_u$ and $h_u$. The router maps $h_u$ to $c_u$ through Eq.~\ref{eq:coeff}, Eq.~\ref{eq:delta} composes the factors, and Eq.~\ref{eq:setting} adds the resulting update $\Delta W_{u,\ell}=\sum_{k}c_{u,k}\,\Delta W^{(k)}_\ell$ to the frozen base weights. For version transfer, the visit performs no tasks. For domain transfer, the base policy attempts \gram{$n$} tasks from the training split of $u$, and the writer reads their intents and answer-stripped reference procedures.
\gram{Adaptation therefore costs} one scouting visit and one router forward pass. It uses no demonstrations from $u$ and no gradient updates.


\section{Experiments}
\label{sec:experiments}
\filled{We answer four research questions.
\textbf{RQ1 (transfer).} Does \ours{} transfer to unseen domains and versions better than in-context learning and LoRA?
\textbf{RQ2 (retention).} Does it keep its performance on the seen interfaces?
\textbf{RQ3 (design).} Which design choices matter: composing a bank instead of generating weights, the scout, the scouting budget, and the KL anchor?
\textbf{RQ4 (analysis).} What do the bank adapters and the router's mixtures look like?}

\subsection{Experimental Setup}
\label{subsec:ExperimentalSetup}

\paragraph{\gram{Benchmarks and evaluation.}}
We evaluate on (i) TimeWarp (TW) \citep{ishmam2026timewarp}, a containerized web benchmark \gram{with three domains (Wiki, News, Shop), each rendered in six versions}, and (ii) WebArena-Lite-v2 (WAL) \citep{liu2026scalecua}, \gram{an updated version of WebArena-Lite \citep{liu2024visualagentbench}, itself a curated subset of WebArena \citep{zhou2024webarena}, with five domains (Reddit, GitLab, CMS, Map, OSS).} \gram{We evaluate} \emph{domain transfer}, in which the unseen interface is a new domain, \gram{on both benchmarks,} and \emph{version transfer}, in which it is a new version, \gram{on TW, holding out v1 and v5 in turn.} We conduct all the experiments with the BrowserUse harness \citep{browser_use2024}, with non-rationale tokens and memory enabled.

\paragraph{Baselines.}
\gram{We compare \ours{} with three baselines on two open-weight models, Qwen-3.5 4B \citep{qwen3.5} and Llama-3.1 8B \citep{grattafiori2024llama3herdmodels}: the base model zero-shot, in-context learning (ICL) \citep{brown2020language}, and LoRA \citep{hu2022lora}. We evaluate each method in two settings. In the \db{\textbf{U}}nseen setting, the evaluated domain or version is left out of the training data and the in-context examples. In the \dr{\textbf{S}}een setting, it is included. In the seen setting, we report LoRA trained only on the evaluated interface (in), which is a target-trained specialist, and LoRA trained on all interfaces (all). Each number is the average of three runs. \S~\ref{subsec:hparams} gives the training details.}




\begin{table}[t]
  \centering
  \small
  \begin{tabular}{l cl rrrr rrr}
    \toprule
    & & & \multicolumn{4}{c}{\textbf{Domains}} & \multicolumn{3}{c}{\textbf{Versions}} \\
    \cmidrule(lr){4-7} \cmidrule(l){8-10}
    \textbf{Model} & \multicolumn{2}{c}{\textbf{Method}} & Wiki & News & Shop & Avg. & v1 & v5 & Avg. \\
    \midrule
    \multirow{8}{*}[-3pt]{\begin{tabular}{@{}c@{}}Qwen-3.5\\4B\end{tabular}} & \multicolumn{2}{c}{Zero-shot} & 19.35 & 30.30 & 20.99 & 22.92 & 17.48 & 19.09 & 18.28 \\\cmidrule{2-10}
     & \multirow{3}{*}{\color{darkblue}\textbf{U}} & In-context Learning & 18.28 & 40.91 & 20.99 & 25.42 & 19.74 & 20.39 & 20.06 \\
     &  & LoRA & 17.20 & 45.45 & 18.52 & 25.42 & 26.54 & 22.01 & 24.27 \\
     &  & \best{\ours} & \best{24.73} & \best{56.06} & \best{25.93} & \best{33.75} & \best{30.42} & \best{24.60} & \best{27.51} \\\cmidrule{2-10}
     & \multirow{4}{*}{\color{darkred}\textbf{S}} & In-context Learning & 20.43 & 43.94 & 23.46 & 27.92 & 20.39 & 20.06 & 20.23 \\
     &  & LoRA (in) & 27.96 & 48.48 & 28.40 & 33.75 & 27.51 & 23.30 & 25.40 \\
     &  & LoRA (all) & 29.03 & 51.52 & 25.93 & 34.17 & 26.86 & 24.27 & 25.57 \\
     &  & \best{\ours} & \best{34.41} & \best{59.09} & \best{32.10} & \best{40.42} & \best{30.74} & \best{25.24} & \best{28.00} \\
    \midrule
    \multirow{8}{*}[-3pt]{\begin{tabular}{@{}c@{}}Llama-3.1\\8B\end{tabular}} & \multicolumn{2}{c}{Zero-shot} & 4.30 & 6.06 & 12.35 & 7.50 & 5.83 & 7.44 & 6.63 \\\cmidrule{2-10}
     & \multirow{3}{*}{\color{darkblue}\textbf{U}} & In-context Learning & 6.45 & 4.55 & 9.88 & 7.08 & 6.80 & 5.50 & 6.15 \\
     &  & LoRA & 5.38 & 12.12 & 8.64 & 8.33 & 18.77 & 15.21 & 16.99 \\
     &  & \best{\ours} & \best{15.05} & \best{18.18} & \best{14.81} & \best{15.83} & \best{20.39} & \best{17.15} & \best{18.77} \\\cmidrule{2-10}
     & \multirow{4}{*}{\color{darkred}\textbf{S}} & In-context Learning & 6.45 & 6.06 & 14.81 & 9.17 & 7.44 & 7.77 & 7.61 \\
     &  & LoRA (in) & 13.98 & 16.67 & 17.28 & 15.83 & 19.42 & 15.86 & 17.64 \\
     &  & LoRA (all) & 11.83 & 22.73 & 16.05 & 16.25 & 20.06 & 17.80 & 18.93 \\
     &  & \best{\ours} & \best{17.20} & \best{27.27} & \best{19.75} & \best{20.83} & \best{21.36} & \best{20.06} & \best{20.71} \\
    \bottomrule
  \end{tabular}
  \caption{\filled{\textbf{TimeWarp results (RQ1, RQ2).} Success rate (\%) on held-out domains and held-out versions.} U/S: unseen/seen setting. \filled{LoRA (in) is trained on the evaluated interface only, and LoRA (all) on all interfaces. Avg.\ is pooled over all tasks.} Best result per model \filled{and setting} in \textbf{bold}.}
  \label{tab:timeWarpResults}
\end{table}

\definecolor{darkblue}{RGB}{0, 0, 139}
\definecolor{darkred}{RGB}{139, 0, 0}

\begin{table}[t]
  \centering
  \small
  \begin{tabular}{l cl rrrrr r}
    \toprule
    \textbf{Model} & \multicolumn{2}{c}{\textbf{Method}} & \textbf{Reddit} & \textbf{GitLab} & \textbf{CMS} & \textbf{Map} & \textbf{OSS} & \textbf{Overall} \\
    \midrule
    \multirow{8}{*}[-3pt]{\begin{tabular}{@{}c@{}}Qwen-3.5\\4B\end{tabular}} & \multicolumn{2}{c}{Zero-shot} & 19.30 & 22.22 & 22.86 & 14.10 & 26.52 & 21.86 \\\cmidrule{2-9}
     & \multirow{3}{*}{\color{darkblue}\textbf{U}} & In-context Learning & 22.81 & 25.56 & 21.90 & 15.38 & 33.33 & 24.89 \\
     &  & LoRA & 26.32 & 30.00 & 27.62 & 17.95 & 37.88 & 29.22 \\
     &  & \best{\ours} & \best{31.58} & \best{35.55} & \best{30.48} & \best{28.21} & \best{41.67} & \best{34.42} \\\cmidrule{2-9}
     & \multirow{4}{*}{\color{darkred}\textbf{S}} & In-context Learning & 29.82 & 31.11 & 27.62 & 17.95 & 34.09 & 28.79 \\
     &  & LoRA (in) & 45.61 & 35.56 & 35.24 & 25.64 & 43.18 & 37.23 \\
     &  & LoRA (all) & 40.35 & 38.89 & 38.10 & 24.36 & 44.70 & 38.10 \\
     &  & \best{\ours} & \best{47.37} & \best{42.22} & \best{40.95} & \best{32.05} & \best{47.73} & \best{42.42} \\
    \midrule
    \multirow{8}{*}[-3pt]{\begin{tabular}{@{}c@{}}Llama-3.1\\8B\end{tabular}} & \multicolumn{2}{c}{Zero-shot} & 5.26 & 12.22 & 6.67 & 0.00 & 6.06 & 6.28 \\\cmidrule{2-9}
     & \multirow{3}{*}{\color{darkblue}\textbf{U}} & In-context Learning & 1.75 & 10.00 & 3.81 & 0.00 & 6.82 & 4.98 \\
     &  & LoRA & 5.26 & 17.78 & 13.33 & 0.00 & 17.42 & 12.12 \\
     &  & \best{\ours} & \best{12.28} & \best{21.11} & \best{18.10} & \best{7.69} & \best{19.70} & \best{16.67} \\\cmidrule{2-9}
     & \multirow{4}{*}{\color{darkred}\textbf{S}} & In-context Learning & 3.51 & 8.89 & 7.62 & 2.56 & 9.85 & 7.14 \\
     &  & LoRA (in) & 42.11 & 22.22 & 20.00 & 5.13 & 25.76 & 22.29 \\
     &  & LoRA (all) & 36.84 & 25.56 & 21.90 & 6.41 & \best{28.03} & 23.59 \\
     &  & \best{\ours} & \best{45.61} & \best{26.67} & \best{24.76} & \best{10.26} & \best{28.03} & \best{26.19} \\
    \bottomrule
  \end{tabular}
  \caption{\filled{\textbf{WebArena-Lite-v2 results (RQ1, RQ2).} Success rate (\%) on held-out domains.} U/S: unseen/seen setting. Overall is \gram{pooled} over all tasks. \gram{Best result per model and setting in \textbf{bold}.}}
  \label{tab:webArenaResults}
\end{table}

\subsection{Results and Analysis}
\label{subsec:ResultsAndAnalysis}

\begin{figure}[t]
  \centering
  \begin{minipage}[b]{0.55\linewidth}

\colorlet{oursrow}{teal!10}

\centering
\small
\setlength{\tabcolsep}{6pt}
\renewcommand{\arraystretch}{1.15}
\begin{tabular}{l|ccc|c}
\toprule
\multirow{2}{*}[-0.4ex]{\begin{tabular}[c]{@{}l@{}}\textbf{\textsc{Hypernetwork}}\\ \textbf{Design}\end{tabular}}
  & \multicolumn{3}{c|}{\textbf{\textsc{Domain}}}
  & \multirow{2}{*}[-0.4ex]{\textbf{Avg}} \\
\cmidrule(lr){2-4}
  & Wiki & News & Shop & \\
\midrule
\textsc{T2L Recons} (\ref{subsec:t2lReconstructor}) & \phantom{0}8.60 & 24.24 & 18.52 & 16.25 \\
\textsc{T2L SFT} (\ref{subsec:t2lSFT})           & 18.28 & 39.39 & 22.22 & 25.42 \\
\midrule
\textsc{Mixture} (\ref{sec:mixtureHypernetwork})          & \underline{19.35} & 48.48 & \underline{24.69} & \underline{29.17} \\
\midrule
\rowcolor{oursrow}
\quad+\textsc{Scout} \gram{$n{=}1$} & \textbf{24.73} & \underline{56.06} & \textbf{25.93} & \textbf{33.75} \\
\quad+\textsc{Scout} \gram{$n{=}3$} & 18.28 & \textbf{57.58} & 22.22 & 30.42 \\
\quad+\textsc{Scout} \gram{$n{=}5$} & 17.20 & \underline{56.06} & \textbf{25.93} & 30.83 \\
\bottomrule
\end{tabular}
\captionof{table}{\filled{\textbf{Hypernetwork design (RQ3).}} Success rate (\%) on unseen TimeWarp domains with Qwen-3.5 4B. \filled{\textsc{T2L} rows generate every adapter weight, \textsc{Mixture} composes the frozen bank, and +\textsc{Scout} conditions the router on the scout's manual built from $n$ scouted tasks.} Avg is pooled over all tasks. Best result per column in \textbf{bold}, second best \underline{underlined}.}
\label{tab:hypernetworkDesign}

  \end{minipage}\hfill
  \begin{minipage}[b]{0.42\linewidth}
    \centering
    \pgfplotstableread{
Wiki          News          Shop          Avg
0.1827956989  0.5000000000  0.1604938272  0.2625000000
0.2043010753  0.4848484848  0.1728395062  0.2708333333
0.2473118280  0.5606060606  0.2592592593  0.3375000000
0.1935483871  0.5151515152  0.2098765432  0.2875000000
}\klData

\definecolor{cWiki}{HTML}{0072B2}
\definecolor{cNews}{HTML}{E69F00}
\definecolor{cShop}{HTML}{009E73}
\colorlet{cBand}{black!9}

\pgfdeclareplotmark{fstar}{%
  \pgfpathmoveto{\pgfpointpolar{90}{\pgfplotmarksize}}%
  \pgfpathlineto{\pgfpointpolar{126}{0.42\pgfplotmarksize}}%
  \pgfpathlineto{\pgfpointpolar{162}{\pgfplotmarksize}}%
  \pgfpathlineto{\pgfpointpolar{198}{0.42\pgfplotmarksize}}%
  \pgfpathlineto{\pgfpointpolar{234}{\pgfplotmarksize}}%
  \pgfpathlineto{\pgfpointpolar{270}{0.42\pgfplotmarksize}}%
  \pgfpathlineto{\pgfpointpolar{306}{\pgfplotmarksize}}%
  \pgfpathlineto{\pgfpointpolar{342}{0.42\pgfplotmarksize}}%
  \pgfpathlineto{\pgfpointpolar{18}{\pgfplotmarksize}}%
  \pgfpathlineto{\pgfpointpolar{54}{0.42\pgfplotmarksize}}%
  \pgfpathclose\pgfusepathqfillstroke}

\pgfplotsset{
  wiki/.style={color=cWiki, mark=*,         mark size=1.9pt, line width=0.9pt},
  news/.style={color=cNews, mark=square*,   mark size=1.7pt, line width=0.9pt},
  shop/.style={color=cShop, mark=triangle*, mark size=2.4pt, line width=0.9pt},
  avg/.style ={color=black, mark=diamond*,  mark size=2.2pt, line width=1.1pt,
               dashed, mark options={solid}},
  best star/.style={mark=fstar, mark size=4pt,
                    mark options={solid, draw=black, line width=0.35pt}},
  legend image code/.code={\draw[mark repeat=2, mark phase=2, #1]
    plot coordinates {(0cm,0cm) (0.18cm,0cm) (0.36cm,0cm)};},
}

\newcommand{\series}[5]{%
  \addplot[#3, no markers, forget plot]
    table[x expr=\coordindex+1+#5, y expr=100*\thisrow{#2}] {#1};
  \addplot[#3, only marks, forget plot,
           skip coords between index={#4}{\numexpr#4+1\relax}]
    table[x expr=\coordindex+1+#5, y expr=100*\thisrow{#2}] {#1};
  \addplot[#3, best star, only marks, forget plot,
           skip coords between index={0}{#4},
           skip coords between index={\numexpr#4+1\relax}{1000}]
    table[x expr=\coordindex+1+#5, y expr=100*\thisrow{#2}] {#1};
}

\newcommand{\bestband}[1]{%
  \pgfmathsetmacro\bandL{#1+1-0.38}\pgfmathsetmacro\bandR{#1+1+0.38}%
  \begin{pgfonlayer}{axis background}
    \fill[cBand] ({axis cs:\bandL,0}|-{rel axis cs:0,0})
          rectangle ({axis cs:\bandR,0}|-{rel axis cs:0,1});
  \end{pgfonlayer}
}

\newlength{\ydescwidth}\setlength{\ydescwidth}{36.5pt}

\begin{tikzpicture}
\begin{axis}[
  scale only axis, width=\dimexpr\linewidth-\ydescwidth\relax,
  height=2.9cm, set layers,
  xmin=0.5, xmax=4.5, xtick={1,2,3,4},
  xticklabels={$0$, $0.1$, $1.0$, $3.0$},
  xlabel={KL coefficient $\beta$},
  ymin=10, ymax=60, ytick={10,20,...,60},
  ylabel={Score (\%)},
  axis x line*=bottom, axis y line*=left,
  xtick style={draw=none}, tick align=outside,
  ymajorgrids, grid style={black!12, line width=0.4pt},
  tick label style={font=\small}, label style={font=\small},
  name=kl, legend to name=kllegend,
  legend columns=3, legend cell align=left,
  legend style={draw=none, font=\scriptsize, inner sep=0pt,
                /tikz/every even column/.append style={column sep=3pt}},
]
\bestband{2}
\series{\klData}{Wiki}{wiki}{2}{-0.075}
\series{\klData}{News}{news}{2}{-0.025}
\series{\klData}{Shop}{shop}{2}{ 0.025}
\series{\klData}{Avg}{avg}{2}{ 0.075}
\addlegendimage{wiki}\addlegendentry{Wiki}
\addlegendimage{news}\addlegendentry{News}
\addlegendimage{shop}\addlegendentry{Shop}
\addlegendimage{avg}\addlegendentry{Avg (weighted)}
\addlegendimage{only marks, best star, mark size=2.9pt, fill=white}
\addlegendentry{Best per series}
\addlegendimage{legend image code/.code={\fill[cBand] (0cm,-0.07cm) rectangle (0.36cm,0.09cm);}}
\addlegendentry{Best on Avg}
\end{axis}
\node[anchor=south, inner sep=0pt, yshift=3pt] at (kl.outer north)
  {\pgfplotslegendfromname{kllegend}};
\end{tikzpicture}
    \vspace{-0.5cm}
    \captionof{figure}{\filled{\textbf{KL anchor strength (RQ3).} Success rate (\%) on unseen TimeWarp domains with Qwen-3.5 4B as the anchor weight $\beta$ of Eq.~\ref{eq:anchor} varies. The shaded column marks the best average, $\beta=1.0$.}}
    \label{fig:klSweep}
  \end{minipage}
\end{figure}



\paragraph{\gram{RQ1: Transfer to unseen interfaces.}} \gram{Tables~\ref{tab:timeWarpResults} and~\ref{tab:webArenaResults}} report transfer to \db{\textbf{U}}nseen domains and versions on TW and WAL, respectively. \ours~outperforms every baseline on both models and transfer types, often matching the performance of baselines trained on those interfaces. \gram{For example, it matches LoRA (in) on TW domains with both models.} Gains are largest for domain transfer, at 7.5--8.3 \gram{percentage points (pp)} over the best baseline on TW and 3.3--4.6pp on WAL. Version transfer gains are smaller (1.8--3.2pp), \gram{and LoRA is a strong baseline there.} Transfer is strongest for domains that are similar to the training domains. With Qwen-3.5, News gains 26pp over zero-shot, compared to $\sim$5pp on Wiki and Shop. On WAL, OSS gains the most for both models (14--15pp). The one failure case is Map with Llama-3.1, where every unseen method scores zero.
\filled{Composing adapters trained on other interfaces thus recovers most of the benefit of training on the target itself.}

\paragraph{\gram{RQ2: Retention on seen interfaces.}} \ours's transfer gains do not come at the cost of \dr{\textbf{S}}een interface performance.
\gram{In Tables~\ref{tab:timeWarpResults} and~\ref{tab:webArenaResults},} \ours~is the best or tied-best method in every column. It improves over the strongest seen baseline, LoRA (all), by 4.6--6.3pp on TW domains, 1.8--2.4pp on TW versions, and 2.6--4.3pp on WAL. Training LoRA on all domains adds only 0.2--1.3pp over in-domain training, suggesting that \ours's gains do not solely come from more training data.

\paragraph{\gram{RQ3: Design choices.}}
\textit{Hypernetwork design.} \gram{Table~\ref{tab:hypernetworkDesign}} isolates the contribution of each component. \gram{\textsc{T2L Recons} (Appendix~\ref{subsec:t2lReconstructor}), which regresses the stored adapter updates, is the weakest variant. \textsc{T2L SFT} (Appendix~\ref{subsec:t2lSFT}), which trains the same generator through the policy, improves on it by 9.17pp. Replacing weight generation with the mixture hypernetwork (\S\ref{sec:mixtureHypernetwork}) adds a further 3.75pp, mostly from News (+9pp). Conditioning the router on the scout then adds another 4.58pp, for a total gain of 8.33pp over the strongest T2L variant.}
\filled{Composing a frozen bank is easier to learn from a handful of interfaces than generating weights.{ Appendix~\ref{sec:theory:twin} analyzes an idealization of \textsc{T2L Recons}}.}
\textit{\gram{Scouting budget.}} \textsc{Scout} outperforms \textsc{Mixture} on average for every number of \gram{scouted tasks $n$}, but \gram{$n{=}1$} is best. Larger \gram{$n$} gives a marginal gain on News (+1.5pp at \gram{$n{=}3$}) but costs 6--8 points on Wiki, so we use \gram{$n{=}1$} throughout.
\textit{KL regularization.} \filled{Figure~\ref{fig:klSweep} varies the anchor weight $\beta$.} $\beta{=}1.0$ is the best of the tested values on every domain and on average. Removing the KL term ($\beta{=}0$) lowers the average by over 7pp. A stronger penalty ($\beta{=}3.0$) also hurts. \ours~therefore benefits from moderate regularization.

\paragraph{RQ4: Analysis.} The domain adapters of the bank are nearly orthogonal (cosine below 0.16, Figure~\ref{fig:xa_all_kl_cos}), and without the KL anchor they are about twice as similar (Figure~\ref{fig:layerwise}). The router builds a held-out version from its nearest versions, e.g., v1 from v2 and v3 (Figure~\ref{fig:xa_all_kl_comp}). With v6 in the bank, it puts 96\% of the weight on v6, so the mixture weights track interface similarity.



\section{Conclusion}
\label{sec:conclusion}
\filled{Experience with one website makes the next one easier to use, and \ours{} gives small, open-weight web agents the same advantage. A scout writes a short manual of the new interface, and a router, trained only on source interfaces, maps the encoded manual to mixing weights over a frozen adapter bank. The result is a standard LoRA adapter, produced without gradient updates on the target. \ours{} transfers to unseen domains and versions better than in-context learning and LoRA, often matches target-trained specialists. When the web changes, a web agent does not need retraining. It can simply \textit{remix} what it already knows.

}

\bibliography{iclr2027_conference}

@misc{sarch2025vlmagentsgeneratememories,
      title={VLM Agents Generate Their Own Memories: Distilling Experience into Embodied Programs of Thought}, 
      author={Gabriel Sarch and Lawrence Jang and Michael J. Tarr and William W. Cohen and Kenneth Marino and Katerina Fragkiadaki},
      year={2025},
      eprint={2406.14596},
      archivePrefix={arXiv},
      primaryClass={cs.CV},
      url={https://arxiv.org/abs/2406.14596}, 
}

@misc{qu2023hmoehypernetworkbasedmixtureexperts,
      title={HMOE: Hypernetwork-based Mixture of Experts for Domain Generalization}, 
      author={Jingang Qu and Thibault Faney and Ze Wang and Patrick Gallinari and Soleiman Yousef and Jean-Charles de Hemptinne},
      year={2023},
      eprint={2211.08253},
      archivePrefix={arXiv},
      primaryClass={cs.LG},
      url={https://arxiv.org/abs/2211.08253}, 
}

@article{liu2024visualagentbench,
  title={VisualAgentBench: Towards Large Multimodal Models as Visual Foundation Agents},
  author={Liu, Xiao and Zhang, Tianjie and Gu, Yu and Iong, Iat Long and Xu, Yifan and Song, Xixuan and Zhang, Shudan and Lai, Hanyu and Liu, Xinyi and Zhao, Hanlin and others},
  journal={arXiv preprint arXiv:2408.06327},
  year={2024}
}

@misc{chen2024automanualconstructinginstructionmanuals,
      title={AutoManual: Constructing Instruction Manuals by LLM Agents via Interactive Environmental Learning}, 
      author={Minghao Chen and Yihang Li and Yanting Yang and Shiyu Yu and Binbin Lin and Xiaofei He},
      year={2024},
      eprint={2405.16247},
      archivePrefix={arXiv},
      primaryClass={cs.AI},
      url={https://arxiv.org/abs/2405.16247}, 
}

@misc{bai2026stresswebdiagnosticbenchmarkweb,
      title={StressWeb: A Diagnostic Benchmark for Web Agent Robustness under Realistic Interaction Variability}, 
      author={Haoyue Bai and Dong Wang and Long Chen and Bingguang Hao and Pengyang Shao and Yonghui Yang and Yicheng He and Chenyi Zhuang},
      year={2026},
      eprint={2604.16385},
      archivePrefix={arXiv},
      primaryClass={cs.SE},
      url={https://arxiv.org/abs/2604.16385}, 
}

@article{Kara2025WAREXWA,
  title={WAREX: Web Agent Reliability Evaluation on Existing Benchmarks},
  author={Su Kara and Fazle Elahi Faisal and Suman Nath},
  journal={ArXiv},
  year={2025},
  volume={abs/2510.03285},
  url={https://api.semanticscholar.org/CorpusID:281843185}
}

@article{Yadav2024WhatMF,
  title={What Matters for Model Merging at Scale?},
  author={Prateek Yadav and Tu Vu and Jonathan Lai and Alexandra Chronopoulou and Manaal Faruqui and Mohit Bansal and Tsendsuren Munkhdalai},
  journal={ArXiv},
  year={2024},
  volume={abs/2410.03617},
  url={https://api.semanticscholar.org/CorpusID:273162841}
}

@misc{charakorn2026doctoloralearninginstantlyinternalize,
      title={Doc-to-LoRA: Learning to Instantly Internalize Contexts}, 
      author={Rujikorn Charakorn and Edoardo Cetin and Shinnosuke Uesaka and Robert Tjarko Lange},
      year={2026},
      eprint={2602.15902},
      archivePrefix={arXiv},
      primaryClass={cs.CL},
      url={https://arxiv.org/abs/2602.15902}, 
}

@misc{hotsko2026code2lorahypernetworkgeneratedadapterscode,
      title={Code2LoRA: Hypernetwork-Generated Adapters for Code Language Models under Software Evolution}, 
      author={Liliana Hotsko and Yinxi Li and Yuntian Deng and Pengyu Nie},
      year={2026},
      eprint={2606.06492},
      archivePrefix={arXiv},
      primaryClass={cs.SE},
      url={https://arxiv.org/abs/2606.06492}, 
}

@misc{wang2026hylovqadynamichypernetworkgeneratedlowrank,
      title={HyLoVQA: Dynamic Hypernetwork-Generated Low-Rank Adaptation for Continual Visual Question Answering}, 
      author={Yiran Wang and Chenyi Xiong and Ziyue Qin and Miao Zhang and Kui Xiao and Zhifei Li},
      year={2026},
      eprint={2605.22035},
      archivePrefix={arXiv},
      primaryClass={cs.CV},
      url={https://arxiv.org/abs/2605.22035}, 
}

@article{Su2025LearnbyinteractAD,
  title={Learn-by-interact: A Data-Centric Framework for Self-Adaptive Agents in Realistic Environments},
  author={Hongjin Su and Ruoxi Sun and Jinsung Yoon and Pengcheng Yin and Tao Yu and Sercan {\"O}. Arik},
  journal={ArXiv},
  year={2025},
  volume={abs/2501.10893},
  url={https://api.semanticscholar.org/CorpusID:275757520}
}

@article{Xue2025AnIO,
  title={An Illusion of Progress? Assessing the Current State of Web Agents},
  author={Tianci Xue and Weijian Qi and Tianneng Shi and Chan Hee Song and Boyu Gou and Dawn Xiaodong Song and Huan Sun and Yu Su},
  journal={ArXiv},
  year={2025},
  volume={abs/2504.01382},
  url={https://api.semanticscholar.org/CorpusID:277502135}
}

@inproceedings{Sun2024OSGenesisAG,
  title={OS-Genesis: Automating GUI Agent Trajectory Construction via Reverse Task Synthesis},
  author={Qiushi Sun and Kanzhi Cheng and Zichen Ding and Chuanyang Jin and Yian Wang and Fangzhi Xu and Zhenyu Wu and Chengyou Jia and Liheng Chen and Zhoumianze Liu and Ben Kao and Guohao Li and Junxian He and Yu Qiao and Zhiyong Wu},
  booktitle={Annual Meeting of the Association for Computational Linguistics},
  year={2024},
  url={https://api.semanticscholar.org/CorpusID:275119524}
}

@inproceedings{Pahuja2025ExplorerSE,
  title={Explorer: Scaling Exploration-driven Web Trajectory Synthesis for Multimodal Web Agents},
  author={Vardaan Pahuja and Yadong Lu and Corby Rosset and Boyu Gou and Arindam Mitra and Spencer Whitehead and Yu Su and Ahmed Awadallah},
  booktitle={Annual Meeting of the Association for Computational Linguistics},
  year={2025},
  url={https://api.semanticscholar.org/CorpusID:276408442}
}

@article{Zheng2025SkillWeaverWA,
  title={SkillWeaver: Web Agents can Self-Improve by Discovering and Honing Skills},
  author={Boyuan Zheng and Michael Y. Fatemi and Xiaolong Jin and Zora Zhiruo Wang and Apurva Gandhi and Yueqi Song and Yu Gu and Jayanth Srinivasa and Gao-Wen Liu and Graham Neubig and Yu Su},
  journal={ArXiv},
  year={2025},
  volume={abs/2504.07079},
  url={https://api.semanticscholar.org/CorpusID:277634081}
}

@misc{wang2026webxskillskilllearningautonomous,
      title={WebXSkill: Skill Learning for Autonomous Web Agents}, 
      author={Zhaoyang Wang and Qianhui Wu and Xuchao Zhang and Chaoyun Zhang and Wenlin Yao and Fazle Elahi Faisal and Baolin Peng and Si Qin and Suman Nath and Qingwei Lin and Chetan Bansal and Dongmei Zhang and Saravan Rajmohan and Jianfeng Gao and Huaxiu Yao},
      year={2026},
      eprint={2604.13318},
      archivePrefix={arXiv},
      primaryClass={cs.AI},
      url={https://arxiv.org/abs/2604.13318}, 
}

@article{Chen2025GroundedTA,
  title={Grounded Test-Time Adaptation for LLM Agents},
  author={Arthur Haonan Chen and Zuxin Liu and Jianguo Zhang and Akshara Prabhakar and Zhiwei Liu and Shelby Heinecke and Silvio Savarese and Victor Zhong and Caiming Xiong},
  journal={ArXiv},
  year={2025},
  volume={abs/2511.04847},
  url={https://api.semanticscholar.org/CorpusID:286087642}
}

@article{Wang2024AgentWM,
  title={Agent Workflow Memory},
  author={Zora Zhiruo Wang and Jiayuan Mao and Daniel Fried and Graham Neubig},
  journal={ArXiv},
  year={2024},
  volume={abs/2409.07429},
  url={https://api.semanticscholar.org/CorpusID:272592995}
}

@misc{Murty2024NNetNavUL,
  title={NNetNav: Unsupervised Learning of Browser Agents Through Environment Interaction in the Wild},
  author={Shikhar Murty and Dzmitry Bahdanau and Christopher D. Manning},
  year={2024},
  url={https://api.semanticscholar.org/CorpusID:273162280}
}

@inproceedings{
liu2026scalecua,
title={Scale{CUA}: Scaling Open-Source Computer Use Agents with Cross-Platform Data},
author={Zhaoyang Liu and JingJing Xie and Zichen Ding and Zehao Li and Bowen Yang and Zhenyu Wu and Xuehui Wang and Qiushi Sun and Shi Liu and Weiyun Wang and Shenglong Ye and Qingyun Li and Zeyue Tian and Gen Luo and Xiangyu Yue and Biqing Qi and Kai Chen and Bowen Zhou and Yu Qiao and Qifeng Chen and Wenhai Wang},
booktitle={The Fourteenth International Conference on Learning Representations},
year={2026},
url={https://openreview.net/forum?id=yBFUqdJFZn}
}

@inproceedings{brown2020language,
  title     = {Language Models are Few-Shot Learners},
  author    = {Brown, Tom and Mann, Benjamin and Ryder, Nick and Subbiah, Melanie and
               Kaplan, Jared D and Dhariwal, Prafulla and Neelakantan, Arvind and
               Shyam, Pranav and Sastry, Girish and Askell, Amanda and others},
  booktitle = {Advances in Neural Information Processing Systems},
  volume    = {33},
  pages     = {1877--1901},
  year      = {2020}
}

@inproceedings{verma-etal-2025-adaptagent,
    title = "{A}dapt{A}gent: Adapting Multimodal Web Agents with Few-Shot Learning from Human Demonstrations",
    author = "Verma, Gaurav  and
      Kaur, Rachneet  and
      Srishankar, Nishan  and
      Zeng, Zhen  and
      Balch, Tucker  and
      Veloso, Manuela",
    editor = "Che, Wanxiang  and
      Nabende, Joyce  and
      Shutova, Ekaterina  and
      Pilehvar, Mohammad Taher",
    booktitle = "Proceedings of the 63rd Annual Meeting of the Association for Computational Linguistics (Volume 1: Long Papers)",
    month = jul,
    year = "2025",
    address = "Vienna, Austria",
    publisher = "Association for Computational Linguistics",
    url = "https://aclanthology.org/2025.acl-long.1008/",
    doi = "10.18653/v1/2025.acl-long.1008",
    pages = "20635--20651",
    ISBN = "979-8-89176-251-0"
}

@article{Zheng2024GPT4VisionIA,
  title={GPT-4V(ision) is a Generalist Web Agent, if Grounded},
  author={Boyuan Zheng and Boyu Gou and Jihyung Kil and Huan Sun and Yu Su},
  journal={ArXiv},
  year={2024},
  volume={abs/2401.01614},
  url={https://api.semanticscholar.org/CorpusID:266741821}
}

@article{Zhang2025PrivWebUA,
  title={PrivWeb: Unobtrusive and Content-aware Privacy Protection For Web Agents},
  author={Shu-Ning Zhang and Yu-Tong Jiang and Rongjun Ma and Yu-Ting Yang and Mingyao Xu and Zhixin Huang and Xin Yi and Hewu Li},
  journal={Proceedings of the 2026 CHI Conference on Human Factors in Computing Systems},
  year={2025},
  url={https://api.semanticscholar.org/CorpusID:281315205}
}

@article{He2024WebVoyagerBA,
  title={WebVoyager: Building an End-to-End Web Agent with Large Multimodal Models},
  author={Hongliang He and Wenlin Yao and Kaixin Ma and Wenhao Yu and Yong Dai and Hongming Zhang and Zhenzhong Lan and Dong Yu},
  journal={ArXiv},
  year={2024},
  volume={abs/2401.13919},
  url={https://api.semanticscholar.org/CorpusID:267211622}
}

@article{Qi2024WebRLTL,
  title={WebRL: Training LLM Web Agents via Self-Evolving Online Curriculum Reinforcement Learning},
  author={Zehan Qi and Xiao Liu and Iat Long Iong and Hanyu Lai and Xueqiao Sun and Xinyue Yang and Jiadai Sun and Yu Yang and Shuntian Yao and Tianjie Zhang and Wei Xu and Jie Tang and Yuxiao Dong},
  journal={ArXiv},
  year={2024},
  volume={abs/2411.02337},
  url={https://api.semanticscholar.org/CorpusID:273821797}
}

@misc{grattafiori2024llama3herdmodels,
      title={The Llama 3 Herd of Models}, 
      author={Aaron Grattafiori and Abhimanyu Dubey and Abhinav Jauhri and Abhinav Pandey and Abhishek Kadian and Ahmad Al-Dahle and Aiesha Letman and Akhil Mathur and Alan Schelten and Alex Vaughan and Amy Yang and Angela Fan and Anirudh Goyal and Anthony Hartshorn and Aobo Yang and Archi Mitra and Archie Sravankumar and Artem Korenev and Arthur Hinsvark and Arun Rao and Aston Zhang and Aurelien Rodriguez and Austen Gregerson and Ava Spataru and Baptiste Roziere and Bethany Biron and Binh Tang and Bobbie Chern and Charlotte Caucheteux and Chaya Nayak and Chloe Bi and Chris Marra and Chris McConnell and Christian Keller and Christophe Touret and Chunyang Wu and Corinne Wong and Cristian Canton Ferrer and Cyrus Nikolaidis and Damien Allonsius and Daniel Song and Danielle Pintz and Danny Livshits and Danny Wyatt and David Esiobu and Dhruv Choudhary and Dhruv Mahajan and Diego Garcia-Olano and Diego Perino and Dieuwke Hupkes and Egor Lakomkin and Ehab AlBadawy and Elina Lobanova and Emily Dinan and Eric Michael Smith and Filip Radenovic and Francisco Guzmán and Frank Zhang and Gabriel Synnaeve and Gabrielle Lee and Georgia Lewis Anderson and Govind Thattai and Graeme Nail and Gregoire Mialon and Guan Pang and Guillem Cucurell and Hailey Nguyen and Hannah Korevaar and Hu Xu and Hugo Touvron and Iliyan Zarov and Imanol Arrieta Ibarra and Isabel Kloumann and Ishan Misra and Ivan Evtimov and Jack Zhang and Jade Copet and Jaewon Lee and Jan Geffert and Jana Vranes and Jason Park and Jay Mahadeokar and Jeet Shah and Jelmer van der Linde and Jennifer Billock and Jenny Hong and Jenya Lee and Jeremy Fu and Jianfeng Chi and Jianyu Huang and Jiawen Liu and Jie Wang and Jiecao Yu and Joanna Bitton and Joe Spisak and Jongsoo Park and Joseph Rocca and Joshua Johnstun and Joshua Saxe and Junteng Jia and Kalyan Vasuden Alwala and Karthik Prasad and Kartikeya Upasani and Kate Plawiak and Ke Li and Kenneth Heafield and Kevin Stone and Khalid El-Arini and Krithika Iyer and Kshitiz Malik and Kuenley Chiu and Kunal Bhalla and Kushal Lakhotia and Lauren Rantala-Yeary and Laurens van der Maaten and Lawrence Chen and Liang Tan and Liz Jenkins and Louis Martin and Lovish Madaan and Lubo Malo and Lukas Blecher and Lukas Landzaat and Luke de Oliveira and Madeline Muzzi and Mahesh Pasupuleti and Mannat Singh and Manohar Paluri and Marcin Kardas and Maria Tsimpoukelli and Mathew Oldham and Mathieu Rita and Maya Pavlova and Melanie Kambadur and Mike Lewis and Min Si and Mitesh Kumar Singh and Mona Hassan and Naman Goyal and Narjes Torabi and Nikolay Bashlykov and Nikolay Bogoychev and Niladri Chatterji and Ning Zhang and Olivier Duchenne and Onur Çelebi and Patrick Alrassy and Pengchuan Zhang and Pengwei Li and Petar Vasic and Peter Weng and Prajjwal Bhargava and Pratik Dubal and Praveen Krishnan and Punit Singh Koura and Puxin Xu and Qing He and Qingxiao Dong and Ragavan Srinivasan and Raj Ganapathy and Ramon Calderer and Ricardo Silveira Cabral and Robert Stojnic and Roberta Raileanu and Rohan Maheswari and Rohit Girdhar and Rohit Patel and Romain Sauvestre and Ronnie Polidoro and Roshan Sumbaly and Ross Taylor and Ruan Silva and Rui Hou and Rui Wang and Saghar Hosseini and Sahana Chennabasappa and Sanjay Singh and Sean Bell and Seohyun Sonia Kim and Sergey Edunov and Shaoliang Nie and Sharan Narang and Sharath Raparthy and Sheng Shen and Shengye Wan and Shruti Bhosale and Shun Zhang and Simon Vandenhende and Soumya Batra and Spencer Whitman and Sten Sootla and Stephane Collot and Suchin Gururangan and Sydney Borodinsky and Tamar Herman and Tara Fowler and Tarek Sheasha and Thomas Georgiou and Thomas Scialom and Tobias Speckbacher and Todor Mihaylov and Tong Xiao and Ujjwal Karn and Vedanuj Goswami and Vibhor Gupta and Vignesh Ramanathan and Viktor Kerkez and Vincent Gonguet and Virginie Do and Vish Vogeti and Vítor Albiero and Vladan Petrovic and Weiwei Chu and Wenhan Xiong and Wenyin Fu and Whitney Meers and Xavier Martinet and Xiaodong Wang and Xiaofang Wang and Xiaoqing Ellen Tan and Xide Xia and Xinfeng Xie and Xuchao Jia and Xuewei Wang and Yaelle Goldschlag and Yashesh Gaur and Yasmine Babaei and Yi Wen and Yiwen Song and Yuchen Zhang and Yue Li and Yuning Mao and Zacharie Delpierre Coudert and Zheng Yan and Zhengxing Chen and Zoe Papakipos and Aaditya Singh and Aayushi Srivastava and Abha Jain and Adam Kelsey and Adam Shajnfeld and Adithya Gangidi and Adolfo Victoria and Ahuva Goldstand and Ajay Menon and Ajay Sharma and Alex Boesenberg and Alexei Baevski and Allie Feinstein and Amanda Kallet and Amit Sangani and Amos Teo and Anam Yunus and Andrei Lupu and Andres Alvarado and Andrew Caples and Andrew Gu and Andrew Ho and Andrew Poulton and Andrew Ryan and Ankit Ramchandani and Annie Dong and Annie Franco and Anuj Goyal and Aparajita Saraf and Arkabandhu Chowdhury and Ashley Gabriel and Ashwin Bharambe and Assaf Eisenman and Azadeh Yazdan and Beau James and Ben Maurer and Benjamin Leonhardi and Bernie Huang and Beth Loyd and Beto De Paola and Bhargavi Paranjape and Bing Liu and Bo Wu and Boyu Ni and Braden Hancock and Bram Wasti and Brandon Spence and Brani Stojkovic and Brian Gamido and Britt Montalvo and Carl Parker and Carly Burton and Catalina Mejia and Ce Liu and Changhan Wang and Changkyu Kim and Chao Zhou and Chester Hu and Ching-Hsiang Chu and Chris Cai and Chris Tindal and Christoph Feichtenhofer and Cynthia Gao and Damon Civin and Dana Beaty and Daniel Kreymer and Daniel Li and David Adkins and David Xu and Davide Testuggine and Delia David and Devi Parikh and Diana Liskovich and Didem Foss and Dingkang Wang and Duc Le and Dustin Holland and Edward Dowling and Eissa Jamil and Elaine Montgomery and Eleonora Presani and Emily Hahn and Emily Wood and Eric-Tuan Le and Erik Brinkman and Esteban Arcaute and Evan Dunbar and Evan Smothers and Fei Sun and Felix Kreuk and Feng Tian and Filippos Kokkinos and Firat Ozgenel and Francesco Caggioni and Frank Kanayet and Frank Seide and Gabriela Medina Florez and Gabriella Schwarz and Gada Badeer and Georgia Swee and Gil Halpern and Grant Herman and Grigory Sizov and Guangyi and Zhang and Guna Lakshminarayanan and Hakan Inan and Hamid Shojanazeri and Han Zou and Hannah Wang and Hanwen Zha and Haroun Habeeb and Harrison Rudolph and Helen Suk and Henry Aspegren and Hunter Goldman and Hongyuan Zhan and Ibrahim Damlaj and Igor Molybog and Igor Tufanov and Ilias Leontiadis and Irina-Elena Veliche and Itai Gat and Jake Weissman and James Geboski and James Kohli and Janice Lam and Japhet Asher and Jean-Baptiste Gaya and Jeff Marcus and Jeff Tang and Jennifer Chan and Jenny Zhen and Jeremy Reizenstein and Jeremy Teboul and Jessica Zhong and Jian Jin and Jingyi Yang and Joe Cummings and Jon Carvill and Jon Shepard and Jonathan McPhie and Jonathan Torres and Josh Ginsburg and Junjie Wang and Kai Wu and Kam Hou U and Karan Saxena and Kartikay Khandelwal and Katayoun Zand and Kathy Matosich and Kaushik Veeraraghavan and Kelly Michelena and Keqian Li and Kiran Jagadeesh and Kun Huang and Kunal Chawla and Kyle Huang and Lailin Chen and Lakshya Garg and Lavender A and Leandro Silva and Lee Bell and Lei Zhang and Liangpeng Guo and Licheng Yu and Liron Moshkovich and Luca Wehrstedt and Madian Khabsa and Manav Avalani and Manish Bhatt and Martynas Mankus and Matan Hasson and Matthew Lennie and Matthias Reso and Maxim Groshev and Maxim Naumov and Maya Lathi and Meghan Keneally and Miao Liu and Michael L. Seltzer and Michal Valko and Michelle Restrepo and Mihir Patel and Mik Vyatskov and Mikayel Samvelyan and Mike Clark and Mike Macey and Mike Wang and Miquel Jubert Hermoso and Mo Metanat and Mohammad Rastegari and Munish Bansal and Nandhini Santhanam and Natascha Parks and Natasha White and Navyata Bawa and Nayan Singhal and Nick Egebo and Nicolas Usunier and Nikhil Mehta and Nikolay Pavlovich Laptev and Ning Dong and Norman Cheng and Oleg Chernoguz and Olivia Hart and Omkar Salpekar and Ozlem Kalinli and Parkin Kent and Parth Parekh and Paul Saab and Pavan Balaji and Pedro Rittner and Philip Bontrager and Pierre Roux and Piotr Dollar and Polina Zvyagina and Prashant Ratanchandani and Pritish Yuvraj and Qian Liang and Rachad Alao and Rachel Rodriguez and Rafi Ayub and Raghotham Murthy and Raghu Nayani and Rahul Mitra and Rangaprabhu Parthasarathy and Raymond Li and Rebekkah Hogan and Robin Battey and Rocky Wang and Russ Howes and Ruty Rinott and Sachin Mehta and Sachin Siby and Sai Jayesh Bondu and Samyak Datta and Sara Chugh and Sara Hunt and Sargun Dhillon and Sasha Sidorov and Satadru Pan and Saurabh Mahajan and Saurabh Verma and Seiji Yamamoto and Sharadh Ramaswamy and Shaun Lindsay and Shaun Lindsay and Sheng Feng and Shenghao Lin and Shengxin Cindy Zha and Shishir Patil and Shiva Shankar and Shuqiang Zhang and Shuqiang Zhang and Sinong Wang and Sneha Agarwal and Soji Sajuyigbe and Soumith Chintala and Stephanie Max and Stephen Chen and Steve Kehoe and Steve Satterfield and Sudarshan Govindaprasad and Sumit Gupta and Summer Deng and Sungmin Cho and Sunny Virk and Suraj Subramanian and Sy Choudhury and Sydney Goldman and Tal Remez and Tamar Glaser and Tamara Best and Thilo Koehler and Thomas Robinson and Tianhe Li and Tianjun Zhang and Tim Matthews and Timothy Chou and Tzook Shaked and Varun Vontimitta and Victoria Ajayi and Victoria Montanez and Vijai Mohan and Vinay Satish Kumar and Vishal Mangla and Vlad Ionescu and Vlad Poenaru and Vlad Tiberiu Mihailescu and Vladimir Ivanov and Wei Li and Wenchen Wang and Wenwen Jiang and Wes Bouaziz and Will Constable and Xiaocheng Tang and Xiaojian Wu and Xiaolan Wang and Xilun Wu and Xinbo Gao and Yaniv Kleinman and Yanjun Chen and Ye Hu and Ye Jia and Ye Qi and Yenda Li and Yilin Zhang and Ying Zhang and Yossi Adi and Youngjin Nam and Yu and Wang and Yu Zhao and Yuchen Hao and Yundi Qian and Yunlu Li and Yuzi He and Zach Rait and Zachary DeVito and Zef Rosnbrick and Zhaoduo Wen and Zhenyu Yang and Zhiwei Zhao and Zhiyu Ma},
      year={2024},
      eprint={2407.21783},
      archivePrefix={arXiv},
      primaryClass={cs.AI},
      url={https://arxiv.org/abs/2407.21783}, 
}

@misc{qwen3.5,
    title  = {{Qwen3.5}: Towards Native Multimodal Agents},
    author = {{Qwen Team}},
    year   = {2026},
    month  = {February},
    url    = {https://qwen.ai/blog?id=qwen3.5}
}

@misc{anthropic_cua,
	author = {Anthropic},
	title = {Introducing computer use, a new Claude 3.5 Sonnet, and Claude 3.5 Haiku},
	howpublished = {\url{https://www.anthropic.com/news/3-5-models-and-computer-use}},
	year = {2024},
	note = {[Accessed 20-09-2026]},
}

@misc{openai_cua,
	author = {Open AI},
	title = {Computer Using Agent},
	howpublished = {\url{https://openai.com/index/computer-using-agent/}},
	year = {2025},
	note = {[Accessed 20-09-2026]},
}

@misc{browser_use2024,
  author = {Müller, Magnus and Žunič, Gregor},
  title = {Browser Use: Enable AI to control your browser},
  year = {2024},
  publisher = {GitHub},
  url = {https://github.com/browser-use/browser-use}
}

@inproceedings{ishmam2026timewarp,
  title         = {{TimeWarp}: Evaluating Web Agents by Revisiting the Past},
  author        = {Ishmam, Md Farhan and Marino, Kenneth},
  booktitle     = {Fortieth Conference on Neural Information Processing Systems
               Evaluations and Datasets Track},
  year          = {2026},
  note          = {To appear},
  eprint        = {2603.04949},
  archivePrefix = {arXiv},
  primaryClass  = {cs.AI},
  url           = {https://arxiv.org/abs/2603.04949}
}

@article{kaelbling1998pomdp,
  title   = {Planning and Acting in Partially Observable Stochastic Domains},
  author  = {Kaelbling, Leslie Pack and Littman, Michael L. and Cassandra, Anthony R.},
  journal = {Artificial Intelligence},
  volume  = {101},
  number  = {1--2},
  pages   = {99--134},
  year    = {1998}
}

@inproceedings{hu2022lora,
  title     = {{LoRA}: Low-Rank Adaptation of Large Language Models},
  author    = {Hu, Edward J. and Shen, Yelong and Wallis, Phillip and
               Allen-Zhu, Zeyuan and Li, Yuanzhi and Wang, Shean and
               Wang, Lu and Chen, Weizhu},
  booktitle = {International Conference on Learning Representations (ICLR)},
  year      = {2022}
}

@inproceedings{charakorn2025t2l,
  title     = {Text-to-{LoRA}: Instant Transformer Adaption},
  author    = {Charakorn, Rujikorn and Cetin, Edoardo and Tang, Yujin and
               Lange, Robert Tjarko},
  booktitle = {Proceedings of the 42nd International Conference on Machine
               Learning (ICML)},
  year      = {2025}
}

@inproceedings{ha2017hypernetworks,
  title     = {HyperNetworks},
  author    = {Ha, David and Dai, Andrew M. and Le, Quoc V.},
  booktitle = {International Conference on Learning Representations (ICLR)},
  year      = {2017}
}

@inproceedings{deng2023mind2web,
    author = {Deng, Xiang and Gu, Yu and Zheng, Boyuan and Chen, Shijie and Stevens, Samuel and Wang, Boshi and Sun, Huan and Su, Yu},
    title = {MIND2WEB: towards a generalist agent for the web},
    year = {2023},
    publisher = {Curran Associates Inc.},
    address = {Red Hook, NY, USA},
    booktitle = {Proceedings of the 37th International Conference on Neural Information Processing Systems},
    articleno = {1220},
    numpages = {24},
    location = {New Orleans, LA, USA},
    series = {NIPS '23}
}

@inproceedings{zhou2024webarena,
  title={WebArena: A Realistic Web Environment for Building Autonomous Agents},
  author={Zhou, Shuyan and Xu, Frank F and Zhu, Hao and Zhou, Xuhui and Lo, Robert and Sridhar, Abishek and Cheng, Xianyi and Bisk, Yonatan and Fried, Daniel and Alon, Uri and others},
  booktitle={International Conference on Learning Representations (ICLR)},
  year={2024}
}

@inproceedings{drouin2024workarena,
author = {Drouin, Alexandre and Gasse, Maxime and Caccia, Massimo and Laradji, Issam H. and Verme, Manuel Del and Marty, Tom and Vazquez, David and Chapados, Nicolas and Lacoste, Alexandre},
title = {WorkArena: how capable are web agents at solving common knowledge work tasks?},
year = {2024},
publisher = {JMLR.org},
booktitle = {Proceedings of the 41st International Conference on Machine Learning},
articleno = {462},
numpages = {21},
location = {Vienna, Austria},
series = {ICML'24}
}

@inproceedings{wortsman2022soups,
  title = 	 {Model soups: averaging weights of multiple fine-tuned models improves accuracy without increasing inference time},
  author =       {Wortsman, Mitchell and Ilharco, Gabriel and Gadre, Samir Ya and Roelofs, Rebecca and Gontijo-Lopes, Raphael and Morcos, Ari S and Namkoong, Hongseok and Farhadi, Ali and Carmon, Yair and Kornblith, Simon and Schmidt, Ludwig},
  booktitle = 	 {Proceedings of the 39th International Conference on Machine Learning},
  pages = 	 {23965--23998},
  year = 	 {2022},
  editor = 	 {Chaudhuri, Kamalika and Jegelka, Stefanie and Song, Le and Szepesvari, Csaba and Niu, Gang and Sabato, Sivan},
  volume = 	 {162},
  series = 	 {Proceedings of Machine Learning Research},
  month = 	 {17--23 Jul},
  publisher =    {PMLR},
  url = 	 {https://proceedings.mlr.press/v162/wortsman22a.html}
}

@inproceedings{ilharco2023taskarithmetic,
      title={Editing Models with Task Arithmetic}, 
      author={Gabriel Ilharco and Marco Tulio Ribeiro and Mitchell Wortsman and Suchin Gururangan and Ludwig Schmidt and Hannaneh Hajishirzi and Ali Farhadi},
      booktitle={International Conference on Learning Representations (ICLR)},
      year={2023},
      eprint={2212.04089},
      archivePrefix={arXiv},
      primaryClass={cs.LG},
      url={https://arxiv.org/abs/2212.04089}, 
}

@inproceedings{yadav2023ties,
      title={{TIES}-Merging: Resolving Interference When Merging Models},
      author={Prateek Yadav and Derek Tam and Leshem Choshen and Colin Raffel and Mohit Bansal},
      booktitle={Thirty-seventh Conference on Neural Information Processing Systems},
      year={2023},
      url={https://openreview.net/forum?id=xtaX3WyCj1}
}

@inproceedings{huang2024lorahub,
    title={LoraHub: Efficient Cross-Task Generalization via Dynamic LoRA Composition}, 
    author={Chengsong Huang and Qian Liu and Bill Yuchen Lin and Tianyu Pang and Chao Du and Min Lin},
    booktitle={First Conference on Language Modeling (COLM)},
    year={2024},
    eprint={2307.13269},
    archivePrefix={arXiv},
    primaryClass={cs.CL}
}

@inproceedings{wu2024mole,
  title={Mixture of LoRA Experts}, 
  author={Xun Wu and Shaohan Huang and Furu Wei},
  booktitle={International Conference on Learning Representations (ICLR)},
  year={2024},
  eprint={2404.13628},
  archivePrefix={arXiv},
  primaryClass={cs.CL},
  url={https://arxiv.org/abs/2404.13628}, 
}

@inproceedings{ostapenko2024arrow,
    author = {Ostapenko, Oleksiy and Su, Zhan and Ponti, Edoardo Maria and Charlin, Laurent and Le Roux, Nicolas and Caccia, Lucas and Sordoni, Alessandro},
    title = {Towards modular LLMs by building and reusing a library of LoRAs},
    year = {2024},
    publisher = {JMLR.org},
    booktitle = {Proceedings of the 41st International Conference on Machine Learning},
    articleno = {1576},
    numpages = {20},
    location = {Vienna, Austria},
    series = {ICML'24}
}

@inproceedings{muqeeth2024phatgoose,
    author = {Muqeeth, Mohammed and Liu, Haokun and Liu, Yufan and Raffel, Colin},
    title = {Learning to route among specialized experts for zero-shot generalization},
    year = {2024},
    publisher = {JMLR.org},
    booktitle = {Proceedings of the 41st International Conference on Machine Learning},
    articleno = {1496},
    numpages = {18},
    location = {Vienna, Austria},
    series = {ICML'24}
}

@article{liang2025dnd,
  title={Drag-and-Drop LLMs: Zero-Shot Prompt-to-Weights},
  journal={arXiv preprint arXiv:2506.16406},
  author={Zhiyuan Liang and Dongwen Tang and Yuhao Zhou and Xuanlei Zhao and Mingjia Shi and Wangbo Zhao and Zekai Li and Peihao Wang and Konstantin Schürholt and Damian Borth and Michael M. Bronstein and Yang You and Zhangyang Wang and Kai Wang},
  year={2025},
  eprint={2506.16406},
  archivePrefix={arXiv},
  primaryClass={cs.LG},
  url={https://arxiv.org/abs/2506.16406},
}

@misc{schulman2020kl, 
    title={Approximating KL divergence}, 
    url={http://joschu.net/blog/kl-approx.html}, 
    journal={John Schulman}, 
    author={Schulman, John}, 
    year={2020}, 
    month={Mar}
}

@inproceedings{xiao2024cpack,
  title={C-Pack: Packed Resources For General Chinese Embeddings},
  author={Xiao, Shitao and Liu, Zheng and Zhang, Peitian and Muennighoff, Niklas and Lian, Defu and Nie, Jian-Yun},
  booktitle={Proceedings of the 47th International ACM SIGIR Conference on Research and Development in Information Retrieval},
  pages={641--649},
  year={2024}
}
\bibliographystyle{iclr2027_conference}

\appendix

\section{Method Details}
\label{app:method}

\subsection{\gram{Adapter Factorization}}
\label{sec:bg:lora}
The factors of an adapter are not unique. For any invertible $Q\in\mathbb{R}^{r\times r}$,
\begin{equation}
    (QA_\ell,\; B_\ell Q^{-1})
    \quad\text{induces}\quad
    (B_\ell Q^{-1})(QA_\ell) = B_\ell A_\ell,
    \label{eq:bg:gauge}
\end{equation}
so different factor pairs give the same update and the same policy. Two adapters trained on similar data can therefore lie far apart in factor space and still behave alike.
\ours{} combines adapters through their updates $\Delta W_\ell$, which Eq.~\ref{eq:bg:gauge} leaves unchanged. \gram{Every comparison we report is either functional (task success) or computed on the updates $\Delta W_\ell$ (Figures~\ref{fig:xa_all_kl_cos} and~\ref{fig:layerwise}), never on the factors.}

\subsection{KL Anchor}
\label{subsec:klAnchor}
\begin{equation}
\begin{aligned}
    \tilde d^{(k)}_{i}
    &:= \operatorname{clip}\!\left(
    \log \pi_\theta(y_i\mid x,y_{<i})
    - \log \pi^{(k)}(y_i\mid x,y_{<i}),\ \tau_{\min},\ \tau_{\max}
    \right),\\
    \kappa^{(k)}_{i}
    &:= \exp\big(\tilde d^{(k)}_{i}\big)-\tilde d^{(k)}_{i}-1,
\end{aligned}
\label{eq:anchor_penalty}
\end{equation}
where $\tau_{\min}<0<\tau_{\max}$ are clipping bounds\filled{, set to $\tau_{\min}=-20$ and $\tau_{\max}=8$}. This penalty is nonnegative and vanishes when the two policies assign the same probability to the demonstrated token. It has the form of the $k_3$ estimator of the KL divergence \citep{schulman2020kl}, but it is evaluated on demonstrated tokens instead of samples from $\pi^{(k)}$. It therefore anchors the probabilities of those tokens without constraining the full output distribution. Own-corpus examples contribute only the SFT term, and anchor-corpus examples contribute only the anchoring term. We sample the anchor examples from $\mathcal{D}_{\setminus k}$ to match the size of $\mathcal{D}_k$. The objective encourages each adapter to specialize to its own interface with less interference when the adapters are mixed.

\subsection{Router Details}
\label{subsec:routerDetails}
\filled{The router is the two-layer perceptron of Eq.~\ref{eq:coeff} with hidden width $d_h=256$ and a SiLU nonlinearity. It has $d_cd_h+d_h+(d_h+1)K$ parameters, about $0.26$M for $d_c=1024$ and any $K\le5$ used here. Its output layer starts at zero, so the initial mixture is uniform and the initial adapter is the mean of the bank. One coefficient vector is shared across all adapted matrices.}

\subsection{\filled{Training Hyperparameters}}
\label{subsec:hparams}
\filled{Bank adapters use rank $r=16$ and scale $\alpha=32$ on the attention and MLP projections, and the KL anchor uses $\beta=1.0$ (Figure~\ref{fig:klSweep}). The router has hidden width $d_h=256$ and trains for 500 steps with AdamW. The scout's encoder is BGE-large-en-v1.5 \citep{xiao2024cpack} with $d_c=1024$. For domain transfer, the scout attempts $n=1$ task (\S\ref{subsec:ResultsAndAnalysis}, RQ3). Table~\ref{tab:hparams} lists the hyperparameters of every component.}
\begin{table}[h]
  \centering
  \small
  \begin{tabular}{@{}lll@{}}
    \toprule
    \textbf{Component} & \textbf{Setting} & \textbf{Value} \\
    \midrule
    \multirow{6}{*}{Bank adapters} & rank $r$ / scale $\alpha$ / dropout & 16 / 32 / 0 \\
      & adapted matrices $\Omega$ & attention q, k, v, o and MLP gate, up, down \\
      & epochs & 6 \\
      & learning rate and schedule & $10^{-4}$, cosine, 10\% warmup \\
      & effective batch size & 4 (1 per device $\times$ 4 accumulation) \\
      & precision and hardware & bf16, DeepSpeed ZeRO-3, one 80\,GB GPU \\
    \midrule
    \multirow{3}{*}{KL anchor} & weight $\beta$ & 1.0 \\
      & clipping $[\tau_{\min},\tau_{\max}]$ & $[-20,\,8]$ \\
      & anchor examples & as many as own-corpus examples \\
    \midrule
    \multirow{5}{*}{Router} & hidden width $d_h$ & 256 \\
      & optimizer & AdamW, learning rate $10^{-4}$, no weight decay \\
      & schedule & 20 linear warmup steps, then cosine \\
      & steps / batch size / gradient clip & 500 / 16 / 1.0 \\
      & maximum sequence length & 8192 tokens \\
    \midrule
    \multirow{4}{*}{Scout} & explorer budget & 25 pages, link depth 3 \\
      & scouted tasks $n$ (domain transfer) & 1 \\
      & writer & Qwen-3.5 9B, temperature 0 \\
      & encoder & BGE-large-en-v1.5, $d_c=1024$, 512 tokens \\
    \midrule
    \multirow{2}{*}{Agent decoding} & temperature & 0.01 \\
      & maximum new tokens per step & 2048 \\
    \bottomrule
  \end{tabular}
  \caption{\filled{Training and evaluation hyperparameters.}}
  \label{tab:hparams}
\end{table}

\subsection{Scout Details}
\label{subsec:scoutDetails}

\paragraph{Explorer.}
For version transfer, the explorer runs a scripted, task-free, breadth-first crawl of same-origin links from each site's landing page. It visits up to $25$ pages per site at a maximum link depth of $3$. For every page it records the title, headings, link labels and targets, buttons, and form fields. It also performs one search probe, which submits the label of the first suitable link on the landing page to the first text or search field the crawl finds. For domain transfer, the explorer is task-grounded instead. The frozen base policy, without an adapter, attempts $n$ tasks drawn from the training split of the site, and the explorer replays the pages that the policy visited.

\paragraph{Writer.}
A frozen language model writes the manual $m_v$ from the exploration record. The manual follows a fixed schema of seven sections, which cover purpose and top-level navigation, search, page anatomy, lists and pagination, forms and multi-step flows, interaction pitfalls, and visual style. The writer is instructed to use only labels and paths observed during exploration, and to omit comparisons with other interfaces and any assessment of task performance. For version transfer, an interface spans several sites, and its manual concatenates the per-site manuals. For domain transfer, the writer also receives each attempted task's intent and its reference procedure with the final answer removed. The manual then adds a deterministic eighth section that lists, for each attempted task, its intent and the sequence of actions with the visible label of each target.

\paragraph{Encoder.}
A frozen sentence encoder $\psi$ (\texttt{BGE-large-en-v1.5}\gram{~\citep{xiao2024cpack}}) reads the tokens of the manual and returns their mean-pooled representation $\psi(m_v)\in\mathbb{R}^{d_c}$, with $d_c=1024$.



\section{The \tl{} Baseline in Detail}
\label{sec:app:t2l}
\filled{\tl{} \citep{charakorn2025t2l} is a direct hypernetwork. It maps the same conditioning $h_v$ that \ours{} uses to the two LoRA factors of every adapted matrix. We index the adapted matrices by $s\in\Omega$ in this appendix and the next.}


\subsection{\filled{Architecture}}
A query addresses one factor of one \gram{adapted matrix} through three learnable embedding
tables, indexed by the \gram{matrix's} layer index $l_s$, its module type $m_s$, and
the \emph{role} $\rho$ of the requested factor,
\begin{equation}
  z_{s,\rho} \;=\; T_\phi\!\Big(\big[\,h_v;\; E_{\mathrm{lyr}}(l_s);\;
  E_{\mathrm{mod}}(m_s);\; E_{\mathrm{role}}(\rho)\,\big]\Big),
  \qquad \rho\in\{\mathrm{A},\mathrm{B}\},
  \label{eq:t2lquery}
\end{equation}
%
where each embedding has width $32$ and the trunk $T_\phi$ is a linear
projection to width \gram{$d_T=128$} followed by a SiLU nonlinearity and two residual
blocks. Output heads are shared at the coarsest granularity the target shapes
allow. There is one linear head $P_\kappa$ per distinct weight shape
$(d^{\mathrm{in}},d^{\mathrm{out}})$ \gram{realized} in the target set, shared across
layers and across both roles. The head emits
$\max(r\,d^{\mathrm{in}},\,r\,d^{\mathrm{out}})$ values, and the leading block
is reshaped into the requested factor,
\begin{equation}
  A_s \;=\; \mathrm{mat}_{r\times d^{\mathrm{in}}_s}\!\big(P_{\kappa(s)}\,
  z_{s,\mathrm{A}}\big),
  \qquad
  B_s \;=\; \mathrm{mat}_{d^{\mathrm{out}}_s\times r}\!\big(P_{\kappa(s)}\,
  z_{s,\mathrm{B}}\big).
  \label{eq:t2lheads}
\end{equation}
Because layer, module, and role enter as embeddings, all $2|\Omega|$ queries
are batched and the full adapter is emitted in a single forward pass.

\subsection{Reconstructor}
\label{subsec:t2lReconstructor}
\filled{\textsc{T2L Recons} follows the original training recipe of \tl{} \citep{charakorn2025t2l}. It regresses the stored bank adapters: for each source interface $v_k$, it minimizes the $\ell_1$ distance between the generated update $B_sA_s$ and the stored update $\Delta W^{(k)}_s$, scaled by $10^4$, over 16 randomly drawn matrices per step. It trains for 2000 steps with AdamW (learning rate $10^{-4}$, 100 warmup steps, cosine decay, gradient clipping at 1.0) from the initialization described below.}

\subsection{SFT}
\label{subsec:t2lSFT}
\filled{\textsc{T2L SFT} trains the same generator end to end through the frozen policy with the router objective of Eq.~\ref{eq:obj}, so it receives exactly the training signal of \ours{}. It uses the router's optimizer settings (Appendix~\ref{subsec:hparams}). Its output heads start at zero for $B$ and at a LoRA-style random bias for $A$, so training starts at the base policy.}

\subsection{\gram{Initialization at the oracle's scale}}
A generator that writes weights must write them at the right order of
magnitude from step $0$. Our trained per-version adapters have mean update
magnitude $|\Delta W|\approx 5\times10^{-5}$. With every emitted entry i.i.d.\
$\mathcal{N}(0,\sigma_0^2)$, an entry of $B_sA_s$ sums $r$ products of
independent pairs, so its typical scale is $\sqrt{r}\,\sigma_0^2$. This is
\emph{quadratic} in $\sigma_0$, which is why an innocuous-looking
\gram{initialization} can emit updates orders of magnitude too large and swamp the
frozen computation. We therefore set $\sigma_0=(\varepsilon/\sqrt{r})^{1/2}$
with $\varepsilon=10^{-6}$, which places the generated update well below the
oracle scale at \gram{initialization}. It is near a no-op, but without the
exactly-zero factor that would stall the other factor's gradient through the
product. Where \ours{} starts at the mean trained adapter, \tl{} starts from
almost nothing.


\section{Analysis of the Bank and the Router}
\label{app:analysis}
This appendix answers RQ4 of \S\ref{sec:experiments}: what the bank adapters and the router's mixtures look like.

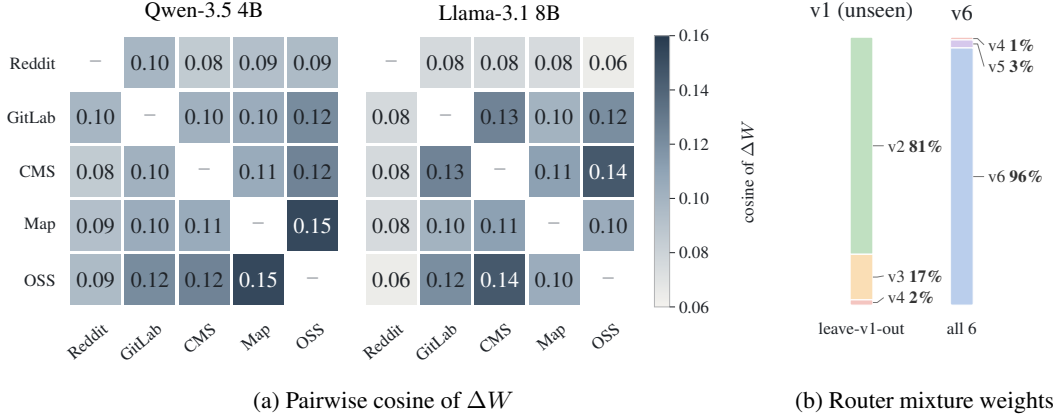
\begin{figure}[t]
    \centering
    \newcommand*{\xafigscale}{0.92}%
    %
    \definecolor{col-v1}{HTML}{1A9E97}
    \definecolor{col-v2}{HTML}{4CA64F}
    \definecolor{col-v3}{HTML}{F0A12B}
    \definecolor{col-v4}{HTML}{E0584B}
    \definecolor{col-v5}{HTML}{8A63C8}
    \definecolor{col-v6}{HTML}{3A74C9}
    \definecolor{vbink}{HTML}{1F2328}%
    \def\vbtint{35}
    %
    \def\vbw{0.42}
    \def\vbtop{-0.03}
    \def\vbbot{-4.97}
    \def\vbround{1.2pt}
    \def\vbgap{0.40}
    \def\vbpad{0.14}
    %
    \newcommand{\vstackbar}[4]{%
      \pgfmathsetmacro\vbx{#1}%
      \pgfmathsetmacro\vbr{\vbx+\vbw}%
      \pgfmathsetmacro\vbc{\vbx+\vbw/2}%
      \node[anchor=south, font=\small, text=vbink] at (\vbc,0.12) {#3};
      \node[anchor=north, font=\scriptsize, text=vbink] at (\vbc,-5.12) {#4};
      \gdef\vbcum{0}\gdef\vbq{0}%
      \begin{scope}
        \clip[rounded corners=\vbround] (\vbx,\vbtop) rectangle (\vbr,\vbbot);
        \foreach \val/\vname in {#2}{%
          \ifdim\val pt>0pt
            \pgfmathsetmacro\vbya{\vbtop-(\vbtop-\vbbot)*\vbcum}%
            \pgfmathsetmacro\vbnew{\vbcum+\val}\global\let\vbcum\vbnew
            \pgfmathsetmacro\vbyb{\vbtop-(\vbtop-\vbbot)*\vbcum}%
            \pgfmathsetmacro\vbym{(\vbya+\vbyb)/2}%
            \pgfmathtruncatemacro\pct{round(100*\val)}%
            \filldraw[fill=col-\vname!\vbtint, draw=white, line width=.5pt]
              (\vbx,\vbya) rectangle (\vbr,\vbyb);
            \pgfmathtruncatemacro\vbqq{\vbq+1}\global\let\vbq\vbqq
            \expandafter\xdef\csname vbM\vbqq\endcsname{\vbym}
            \expandafter\xdef\csname vbY\vbqq\endcsname{\vbym}
            \expandafter\xdef\csname vbN\vbqq\endcsname{\vname}%
            \expandafter\xdef\csname vbP\vbqq\endcsname{\pct}%
          \fi}%
      \end{scope}
      \foreach \vbi in {1,...,\vbq}{%
        \edef\vbyi{\csname vbY\vbi\endcsname}%
        \ifnum\vbi=1
          \pgfmathsetmacro\vbyn{min(\vbyi,\vbtop-\vbpad)}%
        \else
          \pgfmathtruncatemacro\vbj{\vbi-1}%
          \edef\vbyj{\csname vbY\vbj\endcsname}%
          \pgfmathsetmacro\vbyn{min(\vbyi,\vbtop-\vbpad,\vbyj-\vbgap)}%
        \fi
        \expandafter\xdef\csname vbY\vbi\endcsname{\vbyn}}%
      \foreach \vbi in {\vbq,...,1}{%
        \edef\vbyi{\csname vbY\vbi\endcsname}%
        \ifnum\vbi=\vbq
          \pgfmathsetmacro\vbyn{max(\vbyi,\vbbot+\vbpad)}%
        \else
          \pgfmathtruncatemacro\vbj{\vbi+1}%
          \edef\vbyj{\csname vbY\vbj\endcsname}%
          \pgfmathsetmacro\vbyn{max(\vbyi,\vbbot+\vbpad,\vbyj+\vbgap)}%
        \fi
        \expandafter\xdef\csname vbY\vbi\endcsname{\vbyn}}%
      \foreach \vbi in {1,...,\vbq}{%
        \edef\vbm{\csname vbM\vbi\endcsname}%
        \edef\vbyl{\csname vbY\vbi\endcsname}%
        \draw[black!55, line width=.4pt, line join=round]
          (\vbr,\vbm) -- ++(0.08,0) -- (\vbr+0.2,\vbyl);
        \node[font=\scriptsize, text=black!80, anchor=west, inner sep=1.5pt]
          at (\vbr+0.2,\vbyl)
          {\csname vbN\vbi\endcsname~\textbf{\csname vbP\vbi\endcsname\%}};
      }%
    }%
    %
    \begin{subfigure}[t]{0.73\linewidth}
      \centering
      \scalebox{\xafigscale}{%
  \begin{tikzpicture}[x=0.78cm, y=0.78cm, font=\footnotesize]
    \definecolor{hmlo}{HTML}{F4F2EE}   
    \definecolor{hmmid}{HTML}{8FA3B5}   
    \definecolor{hmhi}{HTML}{26394D}   
    \definecolor{hmink}{HTML}{1F2328}
    \definecolor{hmmuted}{HTML}{8A8F96}
    \node[anchor=south, font=\small] at (2.50, 0.12) {Qwen-3.5 4B};
    \node[text=hmmuted] at (0.50,-0.50) {--};
    \fill[hmmid!76!hmlo] (1.03,-0.03) rectangle (1.97,-0.97);  
    \node[text=hmink] at (1.50,-0.50) {0.10};
    \fill[hmmid!50!hmlo] (2.03,-0.03) rectangle (2.97,-0.97);  
    \node[text=hmink] at (2.50,-0.50) {0.08};
    \fill[hmmid!64!hmlo] (3.03,-0.03) rectangle (3.97,-0.97);  
    \node[text=hmink] at (3.50,-0.50) {0.09};
    \fill[hmmid!70!hmlo] (4.03,-0.03) rectangle (4.97,-0.97);  
    \node[text=hmink] at (4.50,-0.50) {0.09};
    \fill[hmmid!76!hmlo] (0.03,-1.03) rectangle (0.97,-1.97);  
    \node[text=hmink] at (0.50,-1.50) {0.10};
    \node[text=hmmuted] at (1.50,-1.50) {--};
    \fill[hmmid!82!hmlo] (2.03,-1.03) rectangle (2.97,-1.97);  
    \node[text=hmink] at (2.50,-1.50) {0.10};
    \fill[hmmid!88!hmlo] (3.03,-1.03) rectangle (3.97,-1.97);  
    \node[text=hmink] at (3.50,-1.50) {0.10};
    \fill[hmhi!22!hmmid] (4.03,-1.03) rectangle (4.97,-1.97);  
    \node[text=hmink] at (4.50,-1.50) {0.12};
    \fill[hmmid!50!hmlo] (0.03,-2.03) rectangle (0.97,-2.97);  
    \node[text=hmink] at (0.50,-2.50) {0.08};
    \fill[hmmid!82!hmlo] (1.03,-2.03) rectangle (1.97,-2.97);  
    \node[text=hmink] at (1.50,-2.50) {0.10};
    \node[text=hmmuted] at (2.50,-2.50) {--};
    \fill[hmmid!90!hmlo] (3.03,-2.03) rectangle (3.97,-2.97);  
    \node[text=hmink] at (3.50,-2.50) {0.11};
    \fill[hmhi!30!hmmid] (4.03,-2.03) rectangle (4.97,-2.97);  
    \node[text=hmink] at (4.50,-2.50) {0.12};
    \fill[hmmid!64!hmlo] (0.03,-3.03) rectangle (0.97,-3.97);  
    \node[text=hmink] at (0.50,-3.50) {0.09};
    \fill[hmmid!88!hmlo] (1.03,-3.03) rectangle (1.97,-3.97);  
    \node[text=hmink] at (1.50,-3.50) {0.10};
    \fill[hmmid!90!hmlo] (2.03,-3.03) rectangle (2.97,-3.97);  
    \node[text=hmink] at (2.50,-3.50) {0.11};
    \node[text=hmmuted] at (3.50,-3.50) {--};
    \fill[hmhi!86!hmmid] (4.03,-3.03) rectangle (4.97,-3.97);  
    \node[text=white] at (4.50,-3.50) {0.15};
    \fill[hmmid!70!hmlo] (0.03,-4.03) rectangle (0.97,-4.97);  
    \node[text=hmink] at (0.50,-4.50) {0.09};
    \fill[hmhi!22!hmmid] (1.03,-4.03) rectangle (1.97,-4.97);  
    \node[text=hmink] at (1.50,-4.50) {0.12};
    \fill[hmhi!30!hmmid] (2.03,-4.03) rectangle (2.97,-4.97);  
    \node[text=hmink] at (2.50,-4.50) {0.12};
    \fill[hmhi!86!hmmid] (3.03,-4.03) rectangle (3.97,-4.97);  
    \node[text=white] at (3.50,-4.50) {0.15};
    \node[text=hmmuted] at (4.50,-4.50) {--};
    \node[rotate=40, anchor=north east, text=hmink, font=\scriptsize] at (0.62,-5.05) {Reddit};
    \node[rotate=40, anchor=north east, text=hmink, font=\scriptsize] at (1.62,-5.05) {GitLab};
    \node[rotate=40, anchor=north east, text=hmink, font=\scriptsize] at (2.62,-5.05) {CMS};
    \node[rotate=40, anchor=north east, text=hmink, font=\scriptsize] at (3.62,-5.05) {Map};
    \node[rotate=40, anchor=north east, text=hmink, font=\scriptsize] at (4.62,-5.05) {OSS};
    \node[anchor=south, font=\small] at (7.95, 0.12) {Llama-3.1 8B};
    \node[text=hmmuted] at (5.95,-0.50) {--};
    \fill[hmmid!32!hmlo] (6.48,-0.03) rectangle (7.42,-0.97);  
    \node[text=hmink] at (6.95,-0.50) {0.08};
    \fill[hmmid!32!hmlo] (7.48,-0.03) rectangle (8.42,-0.97);  
    \node[text=hmink] at (7.95,-0.50) {0.08};
    \fill[hmmid!30!hmlo] (8.48,-0.03) rectangle (9.42,-0.97);  
    \node[text=hmink] at (8.95,-0.50) {0.08};
    \fill[hmmid!8!hmlo] (9.48,-0.03) rectangle (10.42,-0.97);  
    \node[text=hmink] at (9.95,-0.50) {0.06};
    \fill[hmmid!32!hmlo] (5.48,-1.03) rectangle (6.42,-1.97);  
    \node[text=hmink] at (5.95,-1.50) {0.08};
    \node[text=hmmuted] at (6.95,-1.50) {--};
    \fill[hmhi!30!hmmid] (7.48,-1.03) rectangle (8.42,-1.97);  
    \node[text=hmink] at (7.95,-1.50) {0.13};
    \fill[hmmid!80!hmlo] (8.48,-1.03) rectangle (9.42,-1.97);  
    \node[text=hmink] at (8.95,-1.50) {0.10};
    \fill[hmhi!20!hmmid] (9.48,-1.03) rectangle (10.42,-1.97);  
    \node[text=hmink] at (9.95,-1.50) {0.12};
    \fill[hmmid!32!hmlo] (5.48,-2.03) rectangle (6.42,-2.97);  
    \node[text=hmink] at (5.95,-2.50) {0.08};
    \fill[hmhi!30!hmmid] (6.48,-2.03) rectangle (7.42,-2.97);  
    \node[text=hmink] at (6.95,-2.50) {0.13};
    \node[text=hmmuted] at (7.95,-2.50) {--};
    \fill[hmhi!2!hmmid] (8.48,-2.03) rectangle (9.42,-2.97);  
    \node[text=hmink] at (8.95,-2.50) {0.11};
    \fill[hmhi!70!hmmid] (9.48,-2.03) rectangle (10.42,-2.97);  
    \node[text=white] at (9.95,-2.50) {0.14};
    \fill[hmmid!30!hmlo] (5.48,-3.03) rectangle (6.42,-3.97);  
    \node[text=hmink] at (5.95,-3.50) {0.08};
    \fill[hmmid!80!hmlo] (6.48,-3.03) rectangle (7.42,-3.97);  
    \node[text=hmink] at (6.95,-3.50) {0.10};
    \fill[hmhi!2!hmmid] (7.48,-3.03) rectangle (8.42,-3.97);  
    \node[text=hmink] at (7.95,-3.50) {0.11};
    \node[text=hmmuted] at (8.95,-3.50) {--};
    \fill[hmmid!88!hmlo] (9.48,-3.03) rectangle (10.42,-3.97);  
    \node[text=hmink] at (9.95,-3.50) {0.10};
    \fill[hmmid!8!hmlo] (5.48,-4.03) rectangle (6.42,-4.97);  
    \node[text=hmink] at (5.95,-4.50) {0.06};
    \fill[hmhi!20!hmmid] (6.48,-4.03) rectangle (7.42,-4.97);  
    \node[text=hmink] at (6.95,-4.50) {0.12};
    \fill[hmhi!70!hmmid] (7.48,-4.03) rectangle (8.42,-4.97);  
    \node[text=white] at (7.95,-4.50) {0.14};
    \fill[hmmid!88!hmlo] (8.48,-4.03) rectangle (9.42,-4.97);  
    \node[text=hmink] at (8.95,-4.50) {0.10};
    \node[text=hmmuted] at (9.95,-4.50) {--};
    \node[rotate=40, anchor=north east, text=hmink, font=\scriptsize] at (6.07,-5.05) {Reddit};
    \node[rotate=40, anchor=north east, text=hmink, font=\scriptsize] at (7.07,-5.05) {GitLab};
    \node[rotate=40, anchor=north east, text=hmink, font=\scriptsize] at (8.07,-5.05) {CMS};
    \node[rotate=40, anchor=north east, text=hmink, font=\scriptsize] at (9.07,-5.05) {Map};
    \node[rotate=40, anchor=north east, text=hmink, font=\scriptsize] at (10.07,-5.05) {OSS};
    \node[anchor=east, text=hmink, font=\scriptsize] at (-0.08,-0.50) {Reddit};
    \node[anchor=east, text=hmink, font=\scriptsize] at (-0.08,-1.50) {GitLab};
    \node[anchor=east, text=hmink, font=\scriptsize] at (-0.08,-2.50) {CMS};
    \node[anchor=east, text=hmink, font=\scriptsize] at (-0.08,-3.50) {Map};
    \node[anchor=east, text=hmink, font=\scriptsize] at (-0.08,-4.50) {OSS};
    \shade[bottom color=hmlo, top color=hmmid] (10.80,-5) rectangle (11.08,-2.5);
    \shade[bottom color=hmmid, top color=hmhi] (10.80,-2.5) rectangle (11.08,0);
    \draw[hmink!35, line width=0.25pt] (10.80,-5) rectangle (11.08,0);
    \draw[hmink, line width=0.3pt] (11.08,-5.00) -- ++(0.1,0) node[right, font=\scriptsize, inner sep=1.5pt] {0.06};
    \draw[hmink, line width=0.3pt] (11.08,-4.00) -- ++(0.1,0) node[right, font=\scriptsize, inner sep=1.5pt] {0.08};
    \draw[hmink, line width=0.3pt] (11.08,-3.00) -- ++(0.1,0) node[right, font=\scriptsize, inner sep=1.5pt] {0.10};
    \draw[hmink, line width=0.3pt] (11.08,-2.00) -- ++(0.1,0) node[right, font=\scriptsize, inner sep=1.5pt] {0.12};
    \draw[hmink, line width=0.3pt] (11.08,-1.00) -- ++(0.1,0) node[right, font=\scriptsize, inner sep=1.5pt] {0.14};
    \draw[hmink, line width=0.3pt] (11.08,0.00) -- ++(0.1,0) node[right, font=\scriptsize, inner sep=1.5pt] {0.16};
    \node[rotate=90, anchor=north, font=\scriptsize, text=hmink] at (12.25,-2.5) {cosine of $\Delta W$};
  \end{tikzpicture}
      }%
      \caption{Pairwise cosine of $\Delta W$}
      \label{fig:xa_all_kl_cos}
    \end{subfigure}%
    \hfill
    \begin{subfigure}[t]{0.245\linewidth}
      \centering
      \scalebox{\xafigscale}{%
  \begin{tikzpicture}[x=0.78cm, y=0.78cm, font=\footnotesize]
    \vstackbar{0}{0.81/v2, 0.17/v3, 0.02/v4, 0.00/v5, 0.00/v6}
      {v1 (unseen)}{leave-v1-out}
    \vstackbar{1.85}{0.00/v1, 0.00/v2, 0.00/v3, 0.01/v4, 0.03/v5, 0.96/v6}
      {v6}{all 6}
    \coordinate (vbW) at (current bounding box.west);
    \coordinate (vbE) at (current bounding box.east);
    \coordinate (vbT) at (0,0.78);
    \coordinate (vbB) at (0,-6.23);
    \pgfresetboundingbox
    \path[use as bounding box] (vbW |- vbT) rectangle (vbE |- vbB);
  \end{tikzpicture}}%
      \caption{Router mixture weights}
      \label{fig:xa_all_kl_comp}
    \end{subfigure}
    \caption{%
      \textbf{The bank's domain adapters are close to orthogonal, and the router mixes nearby versions (RQ4).}
      \textbf{(a)} Pairwise cosine similarity of the weight updates $\Delta W = \tfrac{\alpha}{r}BA$ between the five
      domain adapters of the KL-anchored WebArena-Lite-v2 bank, computed over the whole model.
      All pairs stay below $0.16$. Reddit is the least aligned with the other domains in both models,
      while OSS overlaps most with Map (Qwen) and CMS (Llama).
      \textbf{(b)} Mixture weights that the router assigns on TimeWarp versions. With v1 held out, it composes v1
      mostly from the adjacent version v2 and then v3. With all six versions in the bank, it puts almost all
      weight on the matching v6 adapter.}
    \label{fig:xa_all_kl}
  \end{figure}

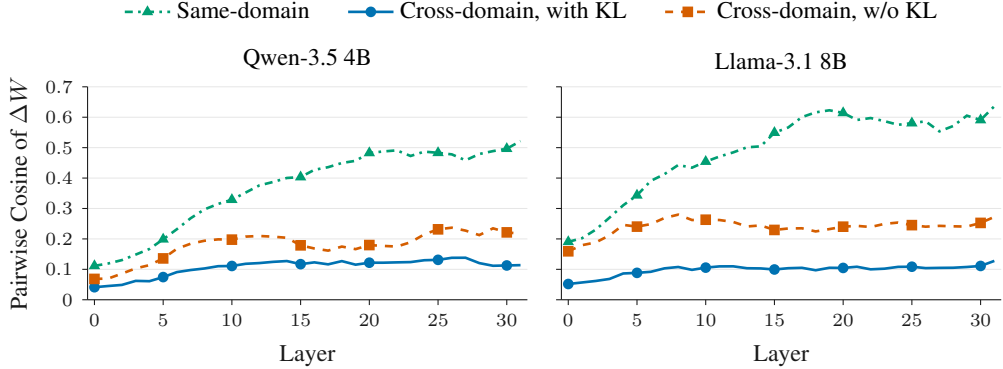
\begin{figure}[t]
  \centering
  \adjustbox{max width=\linewidth}{
\begin{tikzpicture}
  \definecolor{twblue}{HTML}{0072B2}   
  \definecolor{twverm}{HTML}{D55E00}   
  \definecolor{twgreen}{HTML}{009E73}  
  \definecolor{twink}{HTML}{1A1A1A}
  \begin{groupplot}[
    group style={group size=2 by 1, horizontal sep=0.45cm, y descriptions at=edge left},
    width=7.4cm, height=4.4cm,
    xmin=-0.5, xmax=31.5, ymin=0, ymax=0.7,
    xtick={0,5,10,15,20,25,30}, ytick={0,0.1,...,0.7},
    yticklabel style={/pgf/number format/fixed, /pgf/number format/precision=1},
    axis x line*=bottom, axis y line*=left, axis line style={twink, line width=0.4pt}, tick style={twink, line width=0.4pt},
    major tick length=2pt, tick align=outside,
    ymajorgrids, grid style={black!10, line width=0.3pt},
    tick label style={font=\scriptsize, text=twink}, label style={font=\footnotesize, text=twink},
    title style={font=\small, yshift=-3pt},
    xlabel={Layer}, ylabel={Pairwise Cosine of $\Delta W$},
    every axis plot/.append style={line width=1.0pt, mark size=1.5pt, mark repeat=5, mark phase=1,
                                   mark options={solid}},
  ]
  \nextgroupplot[title={Qwen-3.5 4B}, legend to name=lplegend, legend columns=3, legend style={draw=none, font=\footnotesize, /tikz/every even column/.append style={column sep=12pt}}]
    \addplot[twgreen, dashdotted, mark=triangle*] coordinates {(0,0.1114) (1,0.1192) (2,0.1304) (3,0.1484) (4,0.1669) (5,0.1988) (6,0.2320) (7,0.2680) (8,0.2969) (9,0.3151) (10,0.3288) (11,0.3523) (12,0.3759) (13,0.3872) (14,0.4003) (15,0.4032) (16,0.4271) (17,0.4358) (18,0.4495) (19,0.4567) (20,0.4826) (21,0.4874) (22,0.4911) (23,0.4726) (24,0.4869) (25,0.4830) (26,0.4779) (27,0.4584) (28,0.4791) (29,0.4886) (30,0.4965) (31,0.5216)};
    \addlegendentry{Same-domain}
    \addplot[twblue, solid, mark=*] coordinates {(0,0.0412) (1,0.0452) (2,0.0488) (3,0.0620) (4,0.0610) (5,0.0745) (6,0.0909) (7,0.0977) (8,0.1028) (9,0.1103) (10,0.1110) (11,0.1182) (12,0.1206) (13,0.1246) (14,0.1272) (15,0.1170) (16,0.1229) (17,0.1164) (18,0.1270) (19,0.1153) (20,0.1218) (21,0.1218) (22,0.1229) (23,0.1238) (24,0.1299) (25,0.1312) (26,0.1379) (27,0.1381) (28,0.1205) (29,0.1117) (30,0.1127) (31,0.1138)};
    \addlegendentry{Cross-domain, with KL}
    \addplot[twverm, dashed, mark=square*] coordinates {(0,0.0685) (1,0.0697) (2,0.0860) (3,0.1031) (4,0.1138) (5,0.1359) (6,0.1676) (7,0.1842) (8,0.1939) (9,0.1985) (10,0.1976) (11,0.2076) (12,0.2096) (13,0.2074) (14,0.2037) (15,0.1789) (16,0.1685) (17,0.1615) (18,0.1748) (19,0.1664) (20,0.1798) (21,0.1770) (22,0.1749) (23,0.1890) (24,0.2150) (25,0.2315) (26,0.2369) (27,0.2273) (28,0.2125) (29,0.2348) (30,0.2215) (31,0.2171)};
    \addlegendentry{Cross-domain, w/o KL}
  \nextgroupplot[title={Llama-3.1 8B}]
    \addplot[twgreen, dashdotted, mark=triangle*] coordinates {(0,0.1904) (1,0.2023) (2,0.2305) (3,0.2699) (4,0.3106) (5,0.3433) (6,0.3907) (7,0.4129) (8,0.4429) (9,0.4341) (10,0.4541) (11,0.4708) (12,0.4844) (13,0.5011) (14,0.5043) (15,0.5492) (16,0.5662) (17,0.6004) (18,0.6160) (19,0.6228) (20,0.6142) (21,0.5910) (22,0.5973) (23,0.5872) (24,0.5751) (25,0.5805) (26,0.5865) (27,0.5535) (28,0.5710) (29,0.6052) (30,0.5908) (31,0.6361)};
    \addplot[twblue, solid, mark=*] coordinates {(0,0.0518) (1,0.0568) (2,0.0618) (3,0.0684) (4,0.0864) (5,0.0884) (6,0.0919) (7,0.1032) (8,0.1077) (9,0.0983) (10,0.1056) (11,0.1097) (12,0.1099) (13,0.1036) (14,0.1030) (15,0.0995) (16,0.1037) (17,0.1050) (18,0.0970) (19,0.1051) (20,0.1046) (21,0.1085) (22,0.1000) (23,0.1021) (24,0.1081) (25,0.1086) (26,0.1040) (27,0.1048) (28,0.1052) (29,0.1077) (30,0.1111) (31,0.1277)};
    \addplot[twverm, dashed, mark=square*] coordinates {(0,0.1593) (1,0.1794) (2,0.1885) (3,0.2116) (4,0.2461) (5,0.2404) (6,0.2486) (7,0.2688) (8,0.2798) (9,0.2621) (10,0.2631) (11,0.2621) (12,0.2557) (13,0.2409) (14,0.2443) (15,0.2295) (16,0.2343) (17,0.2349) (18,0.2246) (19,0.2323) (20,0.2402) (21,0.2429) (22,0.2391) (23,0.2494) (24,0.2538) (25,0.2453) (26,0.2402) (27,0.2429) (28,0.2414) (29,0.2405) (30,0.2526) (31,0.2723)};
  \end{groupplot}
  \node[anchor=south] at ($(group c1r1.outer north east)!0.5!(group c2r1.outer north west)$) {\pgfplotslegendfromname{lplegend}};
\end{tikzpicture}}
  \caption{\filled{\textbf{The KL anchor keeps adapters of different domains apart (RQ4).} Cosine similarity of the weight updates $\Delta W$ per layer, averaged over pairs of adapters trained on the same domain, on different domains with the KL anchor, and on different domains without it.}}
  \label{fig:layerwise}
\end{figure}

\paragraph{The bank.}
\filled{Figure~\ref{fig:xa_all_kl_cos} shows that the domain adapters of the WAL bank are close to orthogonal. Every pairwise cosine similarity of their updates is below 0.16, so each adapter contributes a distinct direction to the mixture. Figure~\ref{fig:layerwise} measures the same similarity per layer. Adapters trained on the same domain become more similar with depth, reaching a cosine of about 0.5 (Qwen) and 0.6 (Llama) in the last layers. Adapters of different domains stay below 0.14 with the KL anchor, and are about twice as similar without it. The anchor therefore keeps the bank's adapters distinct, which is consistent with its gain \gram{in Figure~\ref{fig:klSweep}}.}

\paragraph{The router.}
\filled{Figure~\ref{fig:xa_all_kl_comp} shows the weights the router assigns on TW versions. With v1 held out, the router builds its adapter mostly from v2 (81\%) and v3 (17\%), so the weights fall with distance from v1 in time. With v6 in the bank, it puts 96\% of the weight on the v6 adapter. The coefficients can thus be read as the model's judgment of which interfaces are similar{. Appendix~\ref{sec:theory:wellposed} shows that, for a linearly independent bank, they are determined by the served adapter}.}

\section{Theoretical Analysis}

\label{sec:theory}

{
We analyze the one design choice that separates \ours{} from direct
generation. The emitted adapter is confined to the convex hull of a frozen
bank, and the interface chooses $K$ coefficients rather than every weight. We
ask what this confinement costs, what it buys, and how direct generation
behaves in a tractable idealization. Full proofs are in
Appendix~\ref{sec:app:proofs}.

\paragraph{Setting.}
\gram{In this appendix, $s\in\Omega$ indexes the adapted matrices (written $\ell$ in \S\ref{sec:methodology}, where $\ell$ is kept free for the logits in Appendix~\ref{sec:app:proofs}), and $p_{\theta,\Delta W}$ is the policy $\pi_{\theta+\Delta\theta}$ written as a function of the update.}
We identify an adapter with its induced update $\Delta W=(\Delta
W_s)_{s\in\Omega}$ in the product space
$\mathcal{W}:=\prod_{s\in\Omega}\mathbb{R}^{d^{\mathrm{out}}_s\times
d^{\mathrm{in}}_s}$, with norm $\|\Delta W\|:=(\sum_{s}\|\Delta
W_s\|_F^2)^{1/2}$ and distance
$\mathrm{dist}(\Delta W,\mathcal{H}):=\min_{W'\in\mathcal{H}}\|\Delta W-W'\|$
to a compact set $\mathcal{H}$. The bank is
$\{\Delta W^{(k)}\}_{k=1}^{K}\subset\mathcal{W}${, $K\ge1$,} with $\Delta
W^{(k)}_s=\tfrac{\alpha}{r}B^{(k)}_sA^{(k)}_s$. For $c\in\Delta^{K-1}${, the probability simplex
$\{c\in\mathbb{R}^K:c_k\ge0,\ \sum_kc_k=1\}$ with vertices $e_1,\dots,e_K$,} the
mixture and its reachable set are
\begin{equation}
  M(c) := \sum_{k=1}^{K} c_k\,\Delta W^{(k)},
  \qquad
  \mathcal{H}_K := M(\Delta^{K-1})
  = \mathrm{conv}\{\Delta W^{(1)},\dots,\Delta W^{(K)}\},
  \label{eq:thm:mix}
\end{equation}
and by Eq.~\ref{eq:exact}, $M(c)$ is exactly the update \ours{} serves under
coefficients $c$. For \gram{an interface} $v$ let $p_{\theta,\Delta W}$ denote the
likelihood of the frozen base with update $\Delta W$ installed, and let
\begin{equation}
  \mathcal{R}_v(\Delta W)
  := \mathbb{E}_{(x,y)\sim\mathcal{D}_v}\big[-\log p_{\theta,\Delta W}(y\mid x)\big]
  \label{eq:thm:risk}
\end{equation}
be the {behavior-cloning} risk of {that interface. On $\mathcal{D}_v$, the token-averaged
objective of Eq.~\ref{eq:obj} equals $\mathcal{R}_v$ times the positive constant
$|\mathcal{D}_v|/N_v$, where $N_v$ is the number of response tokens in $\mathcal{D}_v$, so the
constant changes no minimizer and no result below except the value of the Lipschitz constant
$\Lambda_u$ of Assumption~\ref{ass:lip}.}
$\mathcal{R}_v$ is a finite average of smooth functions of the weights, so it
is continuous on $\mathcal{W}$. 
The held-out {interface} is $u$, {and its oracle $\Delta W^{(u)}$ is an adapter trained on the demonstrations of $u$ itself, the LoRA (in) specialist of \S\ref{sec:experiments}. We write} $\hat c:=c_u$ {for the router output} of Eq.~\ref{eq:coeff} on the held-out embedding. {The
demonstrations $\mathcal{D}_u$ of $u$ are never used by \ours{}; the analysis uses them only to
define $\mathcal{R}_u$ and the oracle.}

\subsection{What the hull can reach}
\label{sec:theory:hull}

\begin{assumption}[Lipschitz risk]\label{ass:lip}
Let $\mathcal{K}_u:=\mathrm{conv}\big(\mathcal{H}_K\cup\{\Delta W^{(u)}\}\big)$.
There is a constant $\Lambda_u<\infty$ such that
$|\mathcal{R}_u(\Delta W)-\mathcal{R}_u(\Delta W')|\le\Lambda_u\|\Delta
W-\Delta W'\|$ for all $\Delta W,\Delta W'\in\mathcal{K}_u$.
\end{assumption}
The set $\mathcal{K}_u$ is \leanfix{convex and} compact, and the base network
is continuously differentiable in its weights, so \leanfix{by the mean value
inequality} the assumption holds with $\Lambda_u$ equal to the largest
gradient norm of $\mathcal{R}_u$ on $\mathcal{K}_u$. The
constant can be large. We use the bound for its direction, not its value.

\begin{theorem}[Hull decomposition of the held-out risk]\label{thm:hull}
Under Assumption~\ref{ass:lip}, a minimizer
$c^\star\in\arg\min_{c\in\Delta^{K-1}}\mathcal{R}_u(M(c))$ exists, and
\begin{equation}
  \mathcal{R}_u(M(\hat c))-\mathcal{R}_u(\Delta W^{(u)})
  \;=\;
  \underbrace{\mathcal{R}_u(M(\hat c))-\mathcal{R}_u(M(c^\star))}_{\text{routing}\;\ge\,0}
  \;+\;
  \underbrace{\mathcal{R}_u(M(c^\star))-\mathcal{R}_u(\Delta W^{(u)})}_{\text{approximation}} ,
  \label{eq:thm:decomp}
\end{equation}
where the approximation term {and the best mixture} satisfy
\begin{equation}
\begin{aligned}
  \mathcal{R}_u(M(c^\star))-\mathcal{R}_u(\Delta W^{(u)})
  &\;\le\; \Lambda_u\,\mathrm{dist}\big(\Delta W^{(u)},\mathcal{H}_K\big), \\
  \mathcal{R}_u(M(c^\star)) &\;\le\; \min_{k\le K}\mathcal{R}_u(\Delta W^{(k)}).
\end{aligned}
  \label{eq:thm:bounds}
\end{equation}
\end{theorem}
\emph{Takeaway: the excess held-out risk of \ours{} splits into a term set by
{the bank, the target and the risk and bounded by the geometry of the bank and the target},
which the head cannot change, and a
term set by the head and the conditioning. The best mixture is never worse
than the best single bank adapter transferred to the held-out {interface}.}
\begin{proof}[Proof sketch]
The decomposition is an identity. For the first bound, {take a point $\Pi$ of the compact set
$\mathcal{H}_K$ nearest to $\Delta W^{(u)}$. It}
is some $M(c_\Pi)$, and Lipschitz continuity on $\mathcal{K}_u$ gives
$\mathcal{R}_u(M(c^\star))\le\mathcal{R}_u(M(c_\Pi))\le\mathcal{R}_u(\Delta
W^{(u)})+\Lambda_u\,\mathrm{dist}(\Delta W^{(u)},\mathcal{H}_K)$. For the
second, every vertex $e_k$ lies in $\Delta^{K-1}$ and $M(e_k)=\Delta W^{(k)}$.
\end{proof}
\begin{remark}[What the theorem predicts]\label{rem:hull}
The theorem is stated in behavior-cloning risk, and the benchmark reports
task success. The bridge assumes the two move together{: success is a nonincreasing function
of risk, so a comparison between two adapters in risk carries over to success}. Three predictions
follow. First, column $u$ of the transfer matrix
\gram{(whose entry $(j,u)$ is the success of the adapter trained on interface $j$ when evaluated on $u$)} lists the performance of every per-{interface}
adapter on $u$\leanfix{, and its off-diagonal entries are exactly the bank
adapters}{ (computed with the bank of the split that holds out $u$, because the anchor corpus
of Eq.~\ref{eq:anchor} depends on the split)}\leanfix{, so the gap between its best off-diagonal
entry and its diagonal entry is an observable ceiling on the approximation term}{, measured in
success}. If \ours{} on $u$
beats every \leanfix{off-diagonal} entry of that column, the head is using
\leanfix{a point of the hull that is not a vertex}. Second, the uniform mixture
$c=\mathbf{1}/K$ is a \leanfix{fixed element} of $\mathcal{H}_K$, so its gap to \ours{}
isolates the routing term's gain over conditioning-blind routing. Third,
$\mathrm{dist}(\Delta W^{(u)},\mathcal{H}_K)$ is computable from the stored
{adapters} alone as a $K$-variable least-squares problem over the simplex. {The theorem
predicts that the best mixture's gap to the oracle, the approximation term, is small when this
distance is small, for comparable $\Lambda_u$; the gap of \ours{} adds the routing term.} The
{versions} are temporally ordered, so the oracle of an interior {version} should lie
closer to the hull of the others than the oracle of a boundary {version}. 
\gram{A leave-one-out panel over all six versions, plotting the {best-mixture gap and the gap
of \ours{}} against the hull distance, would test this.}
\gram{The experiments of \S\ref{sec:experiments} do not include the transfer matrix, the uniform mixture, or the hull-distance panel, so these three predictions remain untested.}
\end{remark}
}

{
\begin{corollary}[Growing the bank]\label{cor:growth}
Let a bank of $K+1$ adapters extend the bank of $K$ by one adapter
$\Delta W^{(K+1)}$, and write $M_K,\mathcal{H}_K$ and
$M_{K+1},\mathcal{H}_{K+1}$ for the two mixtures and reachable sets. Then
$M_{K+1}((c,0))=M_K(c)$ for every $c\in\Delta^{K-1}$, hence
$\mathcal{H}_K\subseteq\mathcal{H}_{K+1}$, and for every {interface} $v$ and every
$\Delta W\in\mathcal{W}$,
\begin{equation}
\begin{aligned}
  \mathrm{dist}(\Delta W,\mathcal{H}_{K+1})&\le\mathrm{dist}(\Delta W,\mathcal{H}_K), \\
  \mathcal{R}^\star_v(K+1)&\le\mathcal{R}^\star_v(K),
  \qquad\text{where}\quad
  \mathcal{R}^\star_v(K):=\min_{c\in\Delta^{K-1}}\mathcal{R}_v(M_K(c)).
\end{aligned}
  \label{eq:cor:growth}
\end{equation}
\end{corollary}
\emph{Takeaway: appending an adapter enlarges what the mixture can reach and
can only lower the best achievable risk on every {interface}, old or new. Any
degradation on an old {interface} after the bank grows is routing error, not lost
capacity.}
\begin{remark}[Continual reading]\label{rem:growth}
{This makes precise what leaving the existing adapters unchanged
(\S\ref{sec:method:training}) guarantees: the best achievable risk on an old interface cannot rise
when the bank grows, while the realized risk can, through the routing term of the retrained
router.} Under sequential arrival, report both
$\mathcal{R}^\star_v(K)$, obtained by minimizing over $c$ on $\mathcal{D}_v$
with the stored bank, and the risk of the realized head.
{Their difference is the routing term, and any rise of the realized risk on an old interface
after the bank grows comes from a rise of this term.} The corollary also predicts a
bank-size curve. With \leanfix{nested} banks of size $2,\dots,5$ drawn from
the {source versions}\leanfix{, each extending the previous one by one adapter},
the best-mixture gap on the held-out {version} is non-increasing in $K$. The
realized gap need not be monotone. {The experiments of \S\ref{sec:experiments} include neither
sequential arrival nor this curve.}
\end{remark}
}

{
\begin{corollary}[Convex twin]\label{cor:jensen}
If $\mathcal{R}_u$ is convex on $\mathcal{K}_u$, then for every
$c\in\Delta^{K-1}$,
\begin{equation}
  \mathcal{R}_u(M(c))\;\le\;\sum_{k=1}^{K}c_k\,\mathcal{R}_u(\Delta W^{(k)}),
  \quad\text{in particular}\quad
  \mathcal{R}_u(M(\mathbf{1}/K))\;\le\;\frac{1}{K}\sum_{k=1}^{K}\mathcal{R}_u(\Delta W^{(k)}).
  \label{eq:cor:jensen}
\end{equation}
\end{corollary}
\emph{Takeaway: in the convex twin, {the held-out risk of any mixture is at most the
corresponding average of the held-out risks of the transferred bank adapters, and the risk of the
uniform mixture is at most their mean.}}
\begin{remark}[Scope of the twin]\label{rem:jensen}
Convexity holds \leanfix{whenever} {the logits are an affine function of the whole update
$\Delta W$, for example when only an untied unembedding matrix is adapted}, because the
negative log-likelihood is convex in the logits\leanfix{; the converse is not
claimed}. {Affinity in each adapted matrix separately is not enough: two adapted matrices in
sequence make the logits bilinear in their updates, and the risk can then fail to be convex.} 
{The adapters of Eq.~\ref{eq:lora} adapt matrices in every layer ($128$ matrices for Qwen-3.5
4B and $224$ for Llama-3.1 8B)}, so for the paper's
configuration the corollary is a simplification and its prediction is
directional. {It predicts that the held-out risk of the uniform mixture on $u$ is at most the
mean held-out risk of the bank adapters. The prediction is about risk. A monotone link carries a
comparison between two adapters to success (Remark~\ref{rem:hull}) but not a comparison with an
average, so the success-level reading (the uniform mixture beats the mean off-diagonal entry of
column $u$ of the transfer matrix) holds if success is, in addition, a concave function of risk.
It fails,
for example, if success behaves like $\exp(-\mathcal{R}_u)$: bank risks $0$ and $10$ and a
uniform-mixture risk of $5$ satisfy the corollary, yet $e^{-5}<(e^{0}+e^{-10})/2$.}
\end{remark}
}

\subsection{Direct generation in a frozen-trunk twin}
\label{sec:theory:twin}

{
The {\textsc{T2L Recons}} baseline trains a trunk and per-shape heads and regresses the
{stored bank} updates. We analyze an idealization that keeps the regression and the heads
and drops the rest.

\begin{assumption}[Frozen-trunk twin]\label{ass:twin}
A direct generator consists of a fixed feature map
$\xi:\mathbb{R}^{d_c}\to\mathbb{R}^{d_\xi}$ and a linear head
$\Gamma\in\mathbb{R}^{D_\Omega\times d_\xi}$ that emits the vectorized update
directly, $\mathrm{vec}\,\Delta W_\Gamma(h)=\Gamma\,\xi(h)$, where
$D_\Omega:=\sum_{s\in\Omega}d^{\mathrm{out}}_sd^{\mathrm{in}}_s$. Write
$\xi_k:=\xi(h_{v_k})$, $\Xi:=[\xi_1,\dots,\xi_K]\in\mathbb{R}^{d_\xi\times K}$,
and $\mathcal{B}:=[\mathrm{vec}\,\Delta W^{(1)},\dots,\mathrm{vec}\,\Delta
W^{(K)}]\in\mathbb{R}^{D_\Omega\times K}$. The features of the {source}
{interfaces} are linearly independent, $\mathrm{rank}\,\Xi=K$, and the head is
trained by gradient descent from $\Gamma=0$ on the squared reconstruction loss
$\sum_{k}\|\Gamma\xi_k-\mathrm{vec}\,\Delta W^{(k)}\|^2$.
\end{assumption}

\begin{proposition}[Direct generation is unnormalized routing]\label{prop:twin}
Under Assumption~\ref{ass:twin}, gradient descent with a sufficiently small
step converges to $\hat\Gamma=\mathcal{B}\,\Xi^{+}$ with
$\Xi^{+}=(\Xi^{\top}\Xi)^{-1}\Xi^{\top}$, the unique minimum-norm interpolant
of the {bank updates}. For every conditioning $h$,
\begin{equation}
  \mathrm{vec}\,\Delta W_{\hat\Gamma}(h)
  \;=\;\mathcal{B}\,\eta(h)
  \;=\;\sum_{k=1}^{K}\eta_k(h)\,\mathrm{vec}\,\Delta W^{(k)},
  \qquad
  \eta(h):=\Xi^{+}\xi(h)\in\mathbb{R}^{K}.
  \label{eq:prop:twin}
\end{equation}
Hence (a) the output at any conditioning lies in the linear span of the bank,
(b) the coefficients $\eta(h)$ depend on the feature geometry
$(\Xi,\xi(h))$ alone and not on the bank or on the policy, and (c) $\eta(h)$
need not lie in $\Delta^{K-1}$. For {any two training features} $\xi_1,\xi_2${, a
conditioning with} $\xi(h)=\xi_1+\xi_2$ {receives} $\Delta W^{(1)}+\Delta W^{(2)}$, twice the
scale of the mean of the two, and {one with} $\xi(h)=\xi_1-\xi_2$ {receives} $\Delta
W^{(1)}-\Delta W^{(2)}$.
\end{proposition}
\emph{Takeaway: in the twin, direct generation at a new interface is routing
over the same bank, with coefficients that are neither normalized nor chosen
through the policy. {Its outputs lie in the span of the bank, so it reaches beyond the hull of
\ours{} only through coefficients outside the simplex, whose scale nothing in its objective
controls (Corollary~\ref{cor:scale}).}}

\begin{corollary}[Scale control]\label{cor:scale}
For every $c\in\Delta^{K-1}$ and every \gram{adapted matrix} $s$,
\begin{equation}
  \|M(c)\|\le\max_{k\le K}\|\Delta W^{(k)}\|
  \qquad\text{and}\qquad
  \|M(c)_s\|_F\le\max_{k\le K}\|\Delta W^{(k)}_s\|_F .
  \label{eq:cor:scale}
\end{equation}
No such bound holds in the twin {over feature vectors, for a nonzero bank}. For any $t>0$ and
any $k$,
{a conditioning with} feature vector $\xi(h)=t\,\xi_k$ yields the output
$t\,\Delta W^{(k)}${, so the scale of the output follows the scale of the features}.
\end{corollary}
\emph{Takeaway: the mixture never emits an update larger than the largest
adapter in the bank, at any conditioning and with no tuning, whereas the
scale of a directly generated update is set by the feature geometry.}

\begin{remark}[Bridge and limits]\label{rem:twin}
\tl{} differs from the twin in {five} ways. It trains the trunk, it emits the
factors $A_s,B_s$ whose product is the update, it uses an $\ell_1$ loss \gram{on the updates}{, it
shares each output head across layers and roles, so its outputs for different matrices are
coupled, and it is trained with AdamW on minibatches}.
{The twin idealizes \textsc{T2L Recons}; \textsc{T2L SFT} replaces the regression by the
objective of Eq.~\ref{eq:obj}, which the twin does not model.}
The twin idealizes the heads, which hold most of \tl{}'s parameters
\gram{(Appendix~\ref{sec:app:t2l})}, and it starts from near zero as \tl{} does
(Appendix~\ref{sec:app:t2l}). It predicts two diagnostics on the generated
held-out adapter. The fraction of $\|\Delta W_{\tl}(h_u)\|^2$ that lies in
$\mathrm{span}\{\Delta W^{(k)}\}$ should be close to one. The scale ratio
$\|\Delta W_{\tl}(h_u)\|/\max_k\|\Delta W^{(k)}\|$ can fall on either side of
one with no mechanism in the objective to correct it, whereas
Corollary~\ref{cor:scale} fixes the ratio at or below one for \ours{}. This is
also why \tl{} required the initialization of Appendix~\ref{sec:app:t2l} and
\ours{} {needs no scale-tuned initialization: no mixture exceeds the scale of the bank}. If the representer step fails for the factored
architecture{ or the shared heads}, the span prediction is the one to test first.
\end{remark}
}

\subsection{Well-posedness of the mixture}
\label{sec:theory:wellposed}

{
\begin{proposition}[Gauge invariance and coefficient identifiability]\label{prop:wellposed}
(a) The objective $\mathcal{L}_{\mathrm{router}}$ of Eq.~\ref{eq:obj} depends on
$F_\phi({E_v})$ only through the induced updates $\{B_sA_s\}_{s\in\Omega}${, because the
policy uses the weights $W_{\theta,s}+\tfrac{\alpha}{r}B_sA_s$}. In
particular, replacing bank member $k$ by
$(Q_kA^{(k)}_s,\,B^{(k)}_sQ_k^{-1})$ for any invertible $Q_k$ leaves $M(c)$
{unchanged for every $c$} and $\mathcal{L}_{\mathrm{router}}$ unchanged {for every $\phi$}.
(b) If $\Delta W^{(1)},\dots,\Delta W^{(K)}$ are linearly independent in
$\mathcal{W}$, the map $c\mapsto M(c)$ is injective on $\Delta^{K-1}$.
(c) {The $\ell_1$ loss on factors} $\|A-A^\ast\|_1+\|B-B^\ast\|_1$ against a
stored {target} $(A^\ast,B^\ast)$ with $A^\ast\neq0$, {is not constant over}
the targets $(QA^\ast,B^\ast Q^{-1})$ {as $Q$ varies}, although all
of them are the same model.
\end{proposition}
\gram{Part (c) is a general remark about objectives on factors. Our \textsc{T2L Recons} baseline does not have this problem, because it regresses the product $B_sA_s$ (Appendix~\ref{subsec:t2lReconstructor}), which is gauge-invariant by Eq.~\ref{eq:bg:gauge}.}

\emph{Takeaway: the training objective sees the model and not its
coordinates, and {for a linearly independent bank} the coefficients \ours{} emits are a
well-defined function of the adapter it serves. A factor-space objective {such as the
$\ell_1$ loss of (c)} is not a function of the target model.}
\begin{remark}[Reading the coefficients]\label{rem:wellposed}
{The hypothesis of part (b) is} checkable from the stored bank as
nonsingularity of the $K\times K$ Gram matrix {$G$, $G_{jk}=\langle\Delta W^{(j)},\Delta
W^{(k)}\rangle$,} of vectorized updates. Under
it, two generated adapters differ if and only if their coefficient vectors
differ, {and their distance is the $G$-weighted distance of the coefficients,
$\|M(c)-M(c')\|^2=(c-c')^{\top}G\,(c-c')$.} This is the sense in which comparisons among generated
adapters are meaningful while comparisons among {the factors of} independently trained adapters
are not ({Appendix}~\ref{sec:bg:lora}). {It makes $c$ a function of the served adapter,
which is a precondition for reading $c$ as the model's judgment of similarity between interfaces,
not a proof of it.}
\end{remark}
}

\section{Proofs}
\label{sec:app:proofs}

{

\subsection{Proof of Theorem~\ref{thm:hull}}
\begin{proof}
\emph{Existence.} $\Delta^{K-1}$ is compact {and nonempty ($K\ge1$)}, $M$ is linear and hence
continuous, and $\mathcal{R}_u$ is continuous on {$\mathcal{H}_K\subseteq\mathcal{K}_u$ by
Assumption~\ref{ass:lip}}, so
$c\mapsto\mathcal{R}_u(M(c))$ is continuous on a compact set and attains its
minimum at some $c^\star$.

\emph{Decomposition.} Eq.~\ref{eq:thm:decomp} follows from adding and
subtracting $\mathcal{R}_u(M(c^\star))$. The routing term is non-negative
because $\hat c\in\Delta^{K-1}$ and $c^\star$ minimizes over $\Delta^{K-1}$.

\emph{First bound.} $\mathcal{H}_K=M(\Delta^{K-1})$ is the image of a
compact convex set under a linear map, so it is compact and convex. The
function $W'\mapsto\|\Delta W^{(u)}-W'\|$ is continuous, so it attains its
minimum on $\mathcal{H}_K$ at some point $\Pi$, and by definition
$\|\Delta W^{(u)}-\Pi\|=\mathrm{dist}(\Delta W^{(u)},\mathcal{H}_K)$. Since
$\Pi\in\mathcal{H}_K$ there is $c_\Pi\in\Delta^{K-1}$ with $\Pi=M(c_\Pi)$.
Both $\Pi$ and $\Delta W^{(u)}$ lie in $\mathcal{K}_u$, so
Assumption~\ref{ass:lip} applies and
\begin{align*}
  \mathcal{R}_u(M(c^\star))
  &\;\le\;\mathcal{R}_u(M(c_\Pi))
  \;=\;\mathcal{R}_u(\Pi) \\
  &\;\le\;\mathcal{R}_u(\Delta W^{(u)})+\Lambda_u\|\Pi-\Delta W^{(u)}\| \\
  &\;=\;\mathcal{R}_u(\Delta W^{(u)})+\Lambda_u\,\mathrm{dist}(\Delta W^{(u)},\mathcal{H}_K).
\end{align*}

\emph{Second bound.} For each $k$, the vertex $e_k$ lies in $\Delta^{K-1}$
and $M(e_k)=\Delta W^{(k)}$, so
$\mathcal{R}_u(M(c^\star))\le\mathcal{R}_u(M(e_k))=\mathcal{R}_u(\Delta
W^{(k)})$. Taking the minimum over $k$ gives Eq.~\ref{eq:thm:bounds}.
\end{proof}

\subsection{Proof of Corollary~\ref{cor:growth}}
\begin{proof}
For $c\in\Delta^{K-1}$, the vector $(c,0)$ lies in $\Delta^{K}$ and
$M_{K+1}((c,0))=\sum_{k\le K}c_k\Delta W^{(k)}+0\cdot\Delta W^{(K+1)}=M_K(c)$.
Hence $\mathcal{H}_K=M_K(\Delta^{K-1})\subseteq M_{K+1}(\Delta^{K})=\mathcal{H}_{K+1}$.
A minimum over a superset is no larger than the minimum over the subset. Applied
to $W'\mapsto\|\Delta W-W'\|$ this gives the distance inequality, and applied
to $\mathcal{R}_v\circ M_{K+1}$ over $\{(c,0):c\in\Delta^{K-1}\}\subseteq\Delta^{K}$
it gives $\mathcal{R}^\star_v(K+1)\le\mathcal{R}^\star_v(K)$. {Both minima exist because
$\mathcal{R}_v$ is continuous and both simplices are compact and nonempty.}
\end{proof}

\subsection{Proof of Corollary~\ref{cor:jensen}}
\begin{proof}
$M(c)=\sum_k c_k\Delta W^{(k)}$ is a convex combination of points of
$\mathcal{K}_u$, and $\mathcal{R}_u$ is convex on $\mathcal{K}_u$, so Jensen's
inequality gives $\mathcal{R}_u(M(c))\le\sum_k c_k\mathcal{R}_u(\Delta
W^{(k)})$. The second statement is the case $c=\mathbf{1}/K$.

For the scope stated in Remark~\ref{rem:jensen}: if {the output logits are an affine function
of $\Delta W$}, the logits for a fixed input {and prefix} are $\ell(\Delta
W)=\ell_0+\mathcal{A}(\Delta W)$ for a fixed vector $\ell_0$ and a fixed
linear map $\mathcal{A}$. The \gram{per-token} negative log-likelihood
$\ell\mapsto\log\sum_j\exp(\ell_j)-\ell_y$ is convex in $\ell$ {(by H\"older's inequality)}, the
composition of a convex function with an affine map is convex, and a {nonnegative combination}
of convex functions is convex {($\mathcal{R}_u$ weights every response token by
$1/|\mathcal{D}_u|$)}. Hence $\mathcal{R}_u$ is convex on all
of $\mathcal{W}$ in that configuration.
\end{proof}

\subsection{Proof of Proposition~\ref{prop:twin}}
\begin{proof}
Write $L(\Gamma):=\|\Gamma\Xi-\mathcal{B}\|_F^2$ for the training loss. Since
$\mathrm{rank}\,\Xi=K\le d_\xi$, the matrix $\Xi^{\top}\Xi\in\mathbb{R}^{K\times
K}$ is invertible and $\Xi^{+}:=(\Xi^{\top}\Xi)^{-1}\Xi^{\top}$ satisfies
$\Xi^{+}\Xi=I_K$. Hence $\hat\Gamma:=\mathcal{B}\Xi^{+}$ obeys
$\hat\Gamma\Xi=\mathcal{B}$ and $L(\hat\Gamma)=0$, so $\hat\Gamma$ is a
minimizer and the minimizer set is the affine space
$\{\Gamma:\Gamma\Xi=\mathcal{B}\}=\hat\Gamma+\{N:N\Xi=0\}$.

\emph{Minimum norm.} Every row of $\hat\Gamma=\mathcal{B}\Xi^{+}$ is a linear
combination of the rows of $\Xi^{\top}$, that is, of the vectors
$\xi_1^{\top},\dots,\xi_K^{\top}$. Every row of an $N$ with $N\Xi=0$ is
orthogonal to each $\xi_k$. Therefore
$\langle\hat\Gamma,N\rangle_F=0$ and
$\|\hat\Gamma+N\|_F^2=\|\hat\Gamma\|_F^2+\|N\|_F^2\ge\|\hat\Gamma\|_F^2$,
with equality only at $N=0$. So $\hat\Gamma$ is the unique minimum-norm
minimizer.

\emph{Gradient descent.} $\nabla L(\Gamma)=2(\Gamma\Xi-\mathcal{B})\Xi^{\top}$. {Let the step
be $\gamma$ with $0<\gamma<1/\lambda_{\max}$, where $\lambda_{\max}:=\lambda_{\max}(\Xi\Xi^{\top})$,
let $\Gamma_{t+1}=\Gamma_t-\gamma\nabla L(\Gamma_t)$ with $\Gamma_0=0$, and let
$X_t:=\Gamma_t-\hat\Gamma$. Since $\hat\Gamma\Xi=\mathcal{B}$, the error obeys
$X_{t+1}=X_t(I-2\gamma\,\Xi\Xi^{\top})$.
(i) The rows of $X_t$ stay in $\mathrm{span}\{\xi_1,\dots,\xi_K\}$, that is, $X_tP=X_t$ for the
projector $P:=\Xi\Xi^{+}$. This holds for $X_0=-\hat\Gamma$ because $\hat\Gamma P=\hat\Gamma$, and
it is preserved because $\Xi\Xi^{\top}P=\Xi\Xi^{\top}$.
(ii) Each step decreases the error. Expanding the square, and using
$\langle X,X\,\Xi\Xi^{\top}\rangle_F=\|X\Xi\|_F^2$ and
$\|Z\Xi^{\top}\|_F^2\le\lambda_{\max}\|Z\|_F^2$ (the matrices $\Xi^{\top}\Xi$ and $\Xi\Xi^{\top}$ have
the same nonzero eigenvalues),
\begin{equation*}
  \|X_{t+1}\|_F^2
  =\|X_t\|_F^2-4\gamma\|X_t\Xi\|_F^2+4\gamma^2\|X_t\Xi\Xi^{\top}\|_F^2
  \le\|X_t\|_F^2-4\gamma\,(1-\gamma\lambda_{\max})\,\|X_t\Xi\|_F^2 .
\end{equation*}
(iii) Summing (ii) over $t$ gives
$\sum_t\|X_t\Xi\|_F^2\le\|X_0\|_F^2/\big(4\gamma(1-\gamma\lambda_{\max})\big)<\infty$, so
$X_t\Xi\to0$.
(iv) By (i), $X_t=X_tP=(X_t\Xi)\,\Xi^{+}\to0$, so $\Gamma_t\to\hat\Gamma$.}

\emph{Output.} For any $h$,
$\mathrm{vec}\,\Delta W_{\hat\Gamma}(h)=\hat\Gamma\xi(h)=\mathcal{B}\,\Xi^{+}\xi(h)=\mathcal{B}\,\eta(h)$,
which is Eq.~\ref{eq:prop:twin}. Statements (a) and (b) read off this
expression. For (c), {$\Xi^{+}\Xi=I_K$ gives $\eta(\xi_k)=\Xi^{+}\Xi e_k=e_k$ for every $k$, so
by linearity $\eta(\xi_1\pm\xi_2)=e_1\pm e_2$, and the outputs are
$\mathcal{B}(e_1\pm e_2)=\mathrm{vec}\,\Delta W^{(1)}\pm\mathrm{vec}\,\Delta W^{(2)}$. The
coordinates of $e_1+e_2$ sum to $2$ and $e_1-e_2$ has a negative coordinate, so neither lies in
$\Delta^{K-1}$; orthonormality is not needed.}
\end{proof}

\subsection{Proof of Corollary~\ref{cor:scale}}
\begin{proof}
For $c\in\Delta^{K-1}$, the triangle inequality and $c_k\ge0$ give
$\|M(c)\|\le\sum_k c_k\|\Delta W^{(k)}\|\le\big(\sum_k c_k\big)\max_k\|\Delta
W^{(k)}\|=\max_k\|\Delta W^{(k)}\|$. The per-\gram{matrix} statement is the same
argument applied to $M(c)_s=\sum_k c_k\Delta W^{(k)}_s$ in the Frobenius norm.
For the twin, $\xi(h)=t\,\xi_k$ gives
$\eta(h)=\Xi^{+}(t\,\xi_k)=t\,\Xi^{+}\Xi e_k=t\,e_k$ and hence
$\mathrm{vec}\,\Delta W_{\hat\Gamma}(h)=t\,\mathrm{vec}\,\Delta W^{(k)}$, whose
norm is $t\|\Delta W^{(k)}\|$ and is unbounded in $t${ when $\Delta W^{(k)}\neq0$}.
\end{proof}

\subsection{Proof of Proposition~\ref{prop:wellposed}}
\begin{proof}
(a) In Eq.~\ref{eq:obj}, $p_{\theta,F_\phi({E_v})}$ is the likelihood of the
network whose weight at \gram{matrix} $s$ is $W_{{\theta,}s}+\tfrac{\alpha}{r}B_sA_s$. Replacing
$(A_s,B_s)$ by $(Q_sA_s,B_sQ_s^{-1})$ leaves this weight, hence the
likelihood and the loss, unchanged. For \ours{}, re-gauging bank member $k$
changes the concatenated factors of Eq.~\ref{eq:delta} but not the product
$\sum_k c_k B^{(k)}_sQ_k^{-1}Q_kA^{(k)}_s=\sum_k c_k B^{(k)}_sA^{(k)}_s$ of
Eq.~\ref{eq:exact}, so $M(c)$ is unchanged for every $c$ and so is
$\mathcal{L}_{\mathrm{router}}$.

(b) If $M(c)=M(c')$ then $\sum_k(c_k-c'_k)\Delta W^{(k)}=0$, and linear
independence forces $c_k=c'_k$ for every $k$.

(c) Fix a generator output $(A,B)$ and take $Q=tI_r$ with $t>0$, so the
re-gauged target is $(tA^\ast,B^\ast/t)$. The loss
$\|A-tA^\ast\|_1+\|B-B^\ast/t\|_1$ is finite at $t=1$ and tends to infinity
as $t\to\infty$ because $A^\ast\neq0$. It is therefore not constant in $t$,
while every $(tA^\ast,B^\ast/t)$ induces the same update $B^\ast A^\ast$.
\end{proof}
}

\section*{Declaration of LLM Usage}

The authors acknowledge the use of LLMs to polish the draft, {to} brainstorm ideas and research directions, to orchestrate experiments, {to} verify every proof in Lean, and {to} review code. We take full responsibility for the paper's content and correctness.

\section*{Reproducibility Statement}

We have taken several steps to ensure our results are reproducible. The complete source code is included in the supplementary material and will be released publicly upon acceptance. It comes with a README describing environment setup, dependencies, and each component. All proofs are formalized in Lean 4, and the formalization is included in the supplementary material{, so} every theorem can be machine-checked. {The source code maps every numbered result of Appendix~\ref{sec:theory} to the Lean declaration that proves it.} All datasets are publicly available and have been cited.

\end{document}